%% file: main.tex
\documentclass[11pt,a4paper,onecolumn]{article}   

\usepackage{arxiv} 
\usepackage[utf8]{inputenc} 
\usepackage[T1]{fontenc}    
\usepackage{url}            
\usepackage{booktabs}       
\usepackage{amsfonts}       
\usepackage{graphicx}
\usepackage{amsmath}
\numberwithin{equation}{section}
\usepackage{amsthm}
\usepackage{amssymb}
\usepackage{float}
\usepackage{framed}
\usepackage{multirow}
\usepackage{tabularx}
\usepackage{marvosym}
\usepackage{setspace}
\usepackage{adjustbox}
\usepackage{threeparttable}
\usepackage{algorithm}
\usepackage{algorithmic}
\usepackage[titletoc,title]{appendix}
\usepackage[british]{babel}
\usepackage[switch]{lineno}
\usepackage{qcircuit}

\usepackage{xcolor}

\usepackage[labelfont=bf]{caption}

\usepackage[colorlinks=true,allcolors={blue}]{hyperref} 

\usepackage{cleveref}
\crefname{subsection}{subsection}{subsections}

\graphicspath{{images/}}

\usepackage{titlesec}
\titleformat{\chapter}{\normalfont\huge}{\thechapter.}{20pt}{\bfseries\huge}

\newcommand{\MODEL}{{MAFER}}
\newcommand{\MODELEXT}{{Multi-resolution Approach to Facial Expression Recognition}}

\newcommand{\affwild}{{Aff-Wild2~\cite{affwild2}}}
\newcommand{\fer}{{FER2013~\cite{fer2013}}}
\newcommand{\raf}{{RAF-DB~\cite{rafdb}}}
\newcommand{\oulu}{{Oulu-CASIA~\cite{oulucasia}}}

\usepackage{xstring}
\makeatletter
\AtBeginDocument{
\let\oldref\ref
\renewcommand{\ref}[1]{\IfBeginWith{#1}{fig:}%
{{\color{blue}Figure~\oldref{#1}}}%
{\IfBeginWith{#1}{tab:}{{\color{blue}Table~\oldref{#1}}}{Unsupported ref start}}}}

\title{MAFER: a Multi-resolution Approach to Facial Expression Recognition}
\author{Fabio Valerio Massoli\thanks{Corresponding author}\\ 
	\texttt{fabio.massoli@isti.cnr.it}\And
    Donato Cafarelli\\
	\texttt{donato.cafarelli@isti.cnr.it} \And 
    Claudio Gennaro\\
	\texttt{claudio.gennaro@isti.cnr.it} \And 
	Giuseppe Amato\\
	\texttt{giuseppe.amato@isti.cnr.it} \And
	Fabrizio Falchi\\
	\texttt{fabrizio.falchi@isti.cnr.it} \\ \\
 	Istituto di Scienza e Tecnologie dell'Informazione ``Alessandro Faedo'' -- CNR, Pisa, Italy\\
}

\begin{document}


\maketitle
\input{sec/0_abstract}


\input{sec/1_introduction}
\input{sec/2_related_works}
\input{sec/3_datasets}
\input{sec/4_training_approach}
\input{sec/5_experimental_results}
\input{sec/6_conclusions}

\subsubsection*{Acknowledgments.} 
We gratefully acknowledge the support of NVIDIA Corporation with the donation of the Titan V GPU used for this research.
This  work  was  partially  supported  by WAC@Lucca funded by Fondazione Cassa di Risparmio di Lucca,
AI4EU - an EC H2020 project (Contract  n.  825619), 
and upon work from COST Action 16101 ``Action MULTI-modal Imaging of FOREnsic SciEnce Evidence (MULTI-FORESEE)'', supported by COST (European Cooperation in Science and Technology). The authors want to thank Daniela Burba for proofreading and correcting the paper.

\bibliographystyle{splncs04}
\bibliography{mybib}

\end{document}

%% file: sec/0_abstract.tex
\begin{abstract}

Emotions play a central role in the social life of every human being, and their study, which represents a multidisciplinary subject, embraces a great variety of research fields, e.g., from psychology to computer science, among others. Especially concerning the latter, the analysis of facial expressions represents a very active research area due to its relevance to human-computer interaction applications, for example. In such a context, Facial Expression Recognition (FER) is the task of recognizing expressions on human faces. Typically, face images are acquired by cameras that have, by nature, different characteristics, such as the output resolution. Moreover, other circumstances might involve cameras placed far from the observed scene, thus obtaining faces with very low resolutions. It has been already shown in the literature that Deep Learning models applied to face recognition experience a degradation in their performance when tested against multi-resolution scenarios if not properly trained. Since the FER task involves analyzing face images that can be acquired with heterogeneous sources, thus involving images with very different quality, it is plausible to expect that resolution plays an important role in such a case too.
Stemming from such a hypothesis, we prove the benefits of multi-resolution training for learning models tasked with recognizing facial expressions. Hence, we propose a two-step learning procedure, named \MODELEXT{} (\MODEL{}), to train Deep Convolutional Neural Networks to empower them to generate robust predictions across a wide range of resolutions. 
A relevant feature of \MODEL{} is that it is task-agnostic, i.e., it can be used complementarily to other objective-related techniques. Specifically, only the first step of the training concerns a multi-resolution environment, while the second one is problem-objective oriented and can be adapted to specific needs. To assess the effectiveness of the proposed approach, we performed an extensive experimental campaign on three different publicly available datasets, namely \fer{}, \raf{}, and \oulu{}. Concerning a multi-resolution context, we observe that with our approach, learning models improve upon the current state-of-the-art while reporting utterly comparable results regarding fix-resolution contexts. Finally, we analyze the performance of our models and observe the higher discrimination power of deep features generated from them.

\end{abstract}

%% file: sec/1_introduction.tex
\section{Introduction} \label{sec:introduction}

\begin{figure*}[t]
\includegraphics[width=\linewidth]{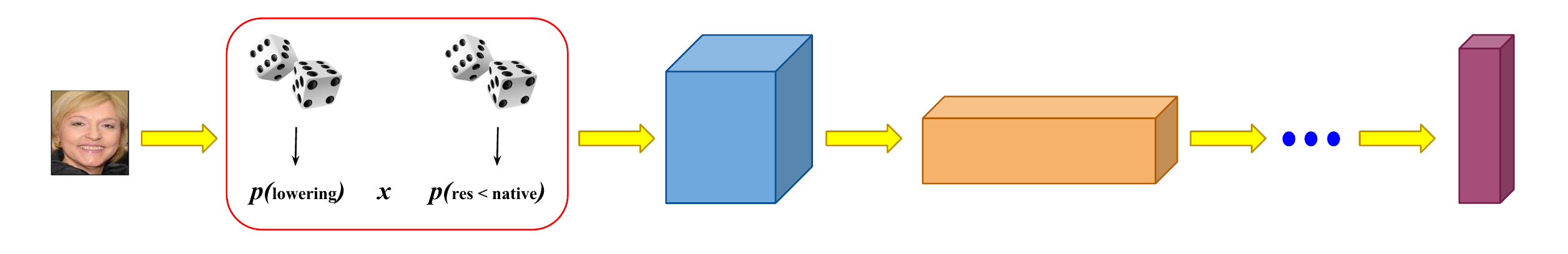}
\caption{Sketch of the multi-resolution training approach we use the pre-train the VGGFace2 model~\cite{cao2017vggface2} on the \affwild{} dataset.}
\label{fig:sketch}
\end{figure*}

In the last two decades, the automatic analysis of facial expressions revealed a very active research area. Especially, due to the key role that affective analysis plays in several research and industrial applications such as human-computer interactions~\cite{kegel2020dynamic,card2018psychology,shi2020human}, customer marketing~\cite{yolcu2020deep,generosi2018deep}, health monitoring~\cite{fei2020deep,sajedi2020uncertainty,muhammad2017facial}, etc. Historically, different formalizations of human expressions have been proposed in the literature. On the one hand, they can be considered as points in a continuous space~\cite{russell1978evidence,russell1980circumplex,whissell1989dictionary,ekman1999basic}, and in such cases, the focus is on arousal and valence. On the other hand, expressions can be interpreted as discrete occurrences that can be used to classify the emotional status of human beings~\cite{ekman1992argument}.

In this manuscript, we focus on the latter interpretation. Specifically, we consider the so-called six universal expressions~\cite{cowie2003describing,dalgleish2000handbook,kollias2018training}, i.e., anger, fear, sadness, happiness, disgust, and surprise.
In the Machine Learning (ML), and especially in the Deep Learning (DL) scientific community, the ability to discern among different facial expressions is a task known as Facial Expression Recognition (FER). In such a context, learning models aim at classifying the human expressions starting from a face image of the given person. 

Face Recognition (FR) is a task that, as FER, involves analyzing facial images. Indeed, even though they have different goals, both tasks aim to classify a person or an expression, starting from the analysis of an image of a human face. Typically, face images are acquired by cameras that have, by their nature, different characteristics such as the output resolution. Moreover, other circumstances might involve cameras placed far from the observed scene, thus returning faces with very low resolution. If such heterogeneity of the input data is not accounted while formulating the training procedure for Deep Convolutional Neural Networks (DCNNs), then it is not guaranteed that the learning models will perform as well as they do in the so-called ``controlled" environment, e.g., when images have high resolutions (a context typically realized in lab experiments). Indeed, it has been already shown in the literature~\cite{massoli2020cross}, that DCNNs suffer from performance degradation when tested against multi-resolution scenarios if not properly trained. Based on such an observation and the similarities between the FER and the FR tasks, we hypothesize that DL models could benefit from multi-resolution training concerning the FER objective too.  

In such a context,  we propose a learning approach, named \MODELEXT{} (\MODEL{}), to train DCNNs to empower them to generate robust predictions across a wide range of resolutions. A relevant feature of \MODEL{} is that it is task-agnostic, i.e., it can be used complementarily to other techniques. 
Specifically, \MODEL{} accounts for a two-step training procedure. A first one, in which the learning model is trained in a multi-resolution environment, and a second one, entirely devoted to the given FER objective at hand. 
To our aim, the first step is the most interesting since it accounts for the multi-resolution technique. Instead, concerning the second step, we use a standard approach to learn a classifier without designing any specific technique, thus emphasizing the real contribution from the multi-resolution training. 

We conceive two flavors for the first training phase. On the one side, we exploit the very same approach presented in~\cite{massoli2020cross} and report the performance gain due to the multi-resolution training compared to a base model. Subsequently, we formulate a lighter version of the mentioned approach, which is less resource-demanding, thus making \MODEL{} suitable to any type of FER-related task. Concerning the last formulation, we report a schematic view in~\autoref{fig:sketch}.

As shown in the red box in~\autoref{fig:sketch}, to simulate a multi-resolution dataset, \MODEL{} exploits two random extractions. The first one is used to decide whether to downsample the image, and the second one to pick the specific resolution. We refer the reader to~\autoref{sec:training_approach} for more details about the multi-resolution training step.


To assess the benefits of our approach, we test the models trained with \MODEL{} against the FER task on three different publicly available datasets,  namely, \fer{}, \raf{}, and \oulu{}. We observe that with our approach, a learning model improves upon the current state-of-the-art when tested against multi-resolution scenarios. Moreover, we investigate the quality of the deep features generated by models trained with \MODEL{} by visually inspecting them using the t-SNE~\cite{maaten2008visualizing} technique and evaluating commonly used metrics in the context of Content-Based Image Retrieval~\cite{donahue2014decaf,class_retrieval} (CBIR).

Regarding the current manuscript, we can summarize our contributions as follows:
\begin{itemize}
\item we propose a multi-resolution procedure, named \MODEL{}, to empower DCNNs to generate robust predictions among a wide range of input face resolution for the FER task;
\item we formulate two flavors concerning the first step of \MODEL{}. On the one side, we exploit the same technique as in~\cite{massoli2020cross}, and on the other, we use that approach as a reference to propose a lighter version for the multi-resolution training step;
\item to assess the benefits of \MODEL{}, we test the performance of learning models on the  \fer{}, \raf{}, and \oulu{} publicly available datasets;
\item we show that with our approach, a learning model improves upon the current state-of-the-art when tested against multi-resolution scenarios while it obtains comparable results on fix-resolution contexts;
\item we analyze the performance of our models and visually compare the quality of the representations 
generated by models trained with and without \MODEL{};
\item we perform CBIR experiments to show the effective discrimination capability of the features generated by models trained with \MODEL{} and compare it to the base model.
\end{itemize}

The remaining part of the manuscript is organized as follows. In \autoref{sec:related_works}, we briefly report works related to our study, while in \autoref{sec:dataset} we describe the datasets we use. In \autoref{sec:training_approach} and \autoref{sec:experimental_results}, we report the training procedure and the experimental results, respectively. Finally, in \autoref{sec:conclusion} we outline our conclusions.

%% file: sec/2_related_works.tex
\section{Related Works} \label{sec:related_works}

In the last two decades, several approaches have been studied to solve the FER task based on different techniques: handcraft features, shallow and deep models. Although each technique came with its pros, generally, the DL-based ones reach the highest performance.

Local Binary Pattern (LBP)~\cite{zhao2011facial,happy2012real}, Gabor wavelets~\cite{bartlett2003real,kotsia2008analysis}, Histogram of Oriented Gradients (HOG)~\cite{zhao2011facial,chen2014facial}, distance and angle relation between landmarks~\cite{michel2003real}, to cite a few, are examples of handcraft-features based approaches. Instead,~\cite{suk2014real,kotsia2006facial,ghimire2013geometric} concerns shallow models based approaches.

Although the mentioned approaches reached modest performance, in recent years, DL-based algorithms have become state-of-the-art to tackle the FER task~\cite{Rouast_2019}.
In 2013,~\cite{tang2013deep} won the ICML face expression recognition challenge~\cite{fer2013} by learning an SVM classifier on top of deep architectures used as backbone features extractors. 
In~\cite{Jung_2015_ICCV}, two CNN are used to detect seven emotions on the Extended Cohn-Kanade~\cite{ck} Dataset (CK+), \oulu{} and MMI~\cite{mmi} datasets. 
In~\cite{hasani2017facial}, the authors use a 3D DCNN followed by a Long Short-Term Memory (LSTM) to analyze and classify facial expressions in videos. 
To overcome the issues related to typical 2D CNN models, Jeong et al.~\cite{jeong2020deep} propose to use a 3D convolution and multiple frames with the normalized LBP feature to extract spatial and temporal features at the same time. 
In~\cite{kollias2017recognition}, the authors propose a CNN-RNN architecture for valence-arousal (VA) recognition on the Aff-Wild database~\cite{zafeiriou2017aff}. 
A Multi-Task learning algorithm is introduced in~\cite{kollias2018multi}, to perform action unit (AU), expression (EX), and Valence Arousal (VA) recognition. In~\cite{kollias2019expression}, the authors propose a multi-task CNN combined with a recurrent neural network for VA and EX recognition.

In~\cite{liu2020effective}, the issue of low-resolution face images is tackled. The authors proposed a novel Hierarchical Convolutional Neural Network (HCNN) that hierarchically assembles shallow CNNs with deep CNNs for effective image super-resolution. Instead,~\cite{shao2021fcnn} proposes an end-to-end network structure for tiny FER, which also exploits super-resolution techniques and encapsulates multiple facial attributes with a feedback mechanism.~\cite{ma2021robust} deals with the problem of facial expression recognition in-the-wild with Convolutional Visual Transformers, while in~\cite{alrubaish2020effects}, the authors investigate the effect of facial expressions on the face biometric systems' reliability. 
A lightweight mobile architecture and a multi-kernel feature facial expression recognition network, which can take into account the speed and accuracy of real-time facial expression recognition is proposed in \cite{li2021efficient}.

In~\cite{conniefer}, the authors address the FER task by proposing a hybrid CNN with a Dense Scale Invariant Feature Transform (Dense SIFT) aggregator model that exploits the combination of deep and handcrafted features to increase the accuracy result on the \fer{} and the CK+~\cite{ck} datasets.
Authors in~\cite{mahmoudiexp2}, propose a FER method based on a CNN model with a specially designed pooling layer, which has learnable weights. Using learnable pooling layers with a Gaussian RBF kernel allowed them to outperform state-of-the-art results on the \raf{} dataset.
The problem of intra-class variations and inter-class similarities in FER datasets is addressed in~\cite{raf-dbresultfarz} with the introduction of a Deep Attentive Center Loss (DACL) method that adaptively selects a subset of significant feature elements for enhanced discrimination.
In~\cite{wang2020regionraf}, the authors propose a Region Attention Network (RAN) to address the problem of occlusion and pose variations in the FER task. This network adaptively captures the importance of facial regions. Furthermore, they annotate several in-the-wild FER datasets with pose and occlusion attributes.
The paper~\cite{florearaf-fb} propose a method called Annealed
Label Transfer for FER task, which makes use of labeled and unlabelled data.
Two FER methods are proposed in~\cite{zhang2020oulu}, namely, Weighted Mixture Deep Convolution Neural Networks (WMDCNN) and deep CNN long short-term memory networks of double-channel weighted mixture (WMCNN-LSTM). The first is based on static images, while the second on image sequences. The WMDCNN network model can provide static facial expressions features to the WMCNN-LSTM network, which exploits them to acquire the temporal features of the image sequence in order to realize the accurate recognition of facial expressions.

%% file: sec/3_datasets.tex
\section{Datasets} \label{sec:dataset}

In this section of the manuscript, we briefly describe the datasets we employ in our study. Specifically, we use the \affwild{} dataset for training purposes only since it is currently used in the ``Affective Behavior Analysis in-the-wild (ABAW) 2020 competition"~\cite{kollias2020analysing} and for that reason, the ground truth for the test set is not available yet.

\subsection{Affect-in-the-Wild 2 (Aff-Wild2)} \label{affwild2dset}

The \affwild{} dataset is the first-ever database annotated for all of the three main behavior tasks: Valence Arousal (VA), Action Unit (AU), and EXpression (EX) classification. Concerning the last task, the dataset consists of 547 videos (collected from YouTube) for a total of $\sim$2.5M frames shared among the seven available expressions: neutral, anger, disgust, fear, happiness, sadness, surprise. Specifically, ``neutral'' means that the specific expression cannot be classified as belonging to any of the remaining available six classes. A pool of seven experts manually annotated frame-by-frame all the available images. The dataset is split into three subsets: training, validation, and test, consisting of 253, 71, and 223 videos, respectively. Moreover, cropped and aligned images are already available in the dataset with a resolution of 112x112 pixels. As mentioned above, since the \affwild{} dataset is currently employed in the ABAW 2020 competition~\cite{kollias2020analysing} the ground truth for the test is not available. Thus, we use this dataset for training purposes only. We refer the reader to \autoref{sec:training_approach} for more details on this topic.

\subsection{Facial Expression Recognition 2013 Dataset (FER2013)}
\begin{figure}[!h]
\includegraphics[width=\linewidth]{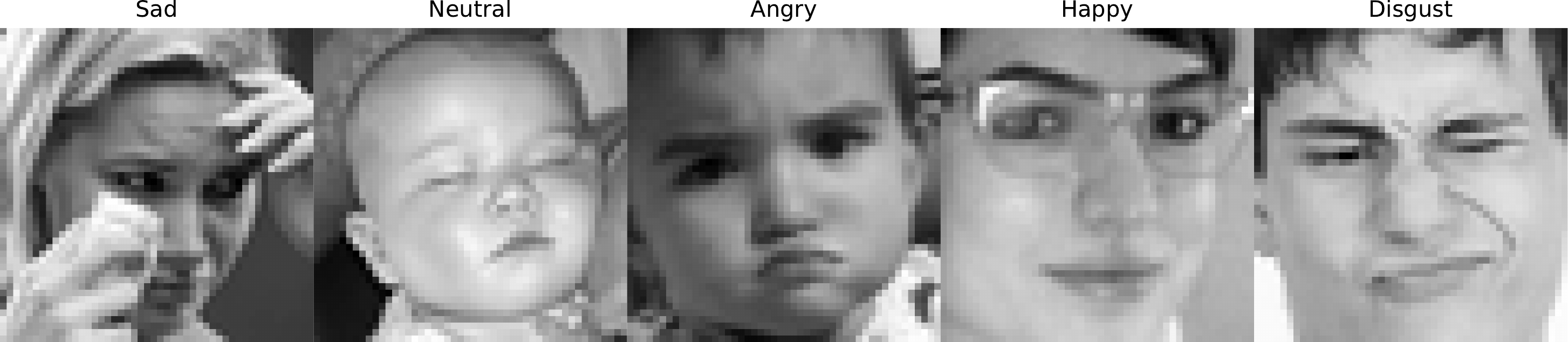}
\caption{Example of face images from the \fer{} dataset. On top of each image, we report the corresponding ground truth expression.}
\label{fig:fer}
\end{figure}

\fer{} is a large-scale and unconstrained database containing 28,709 training, 3,589 validation (public test), and 3,589 (private test) test images belonging to seven different classes: anger, disgust, fear, happiness, sadness, surprise, neutral. Images were collected automatically by the Google image search API, and they are characterized by the same resolution equals to 48x48 pixels. In \autoref{fig:fer}, we report a few examples of facial images from the \fer{} dataset.

\subsection{Real-world Affective Faces (RAF-DB)}
\begin{figure}[!h]
\includegraphics[width=\linewidth]{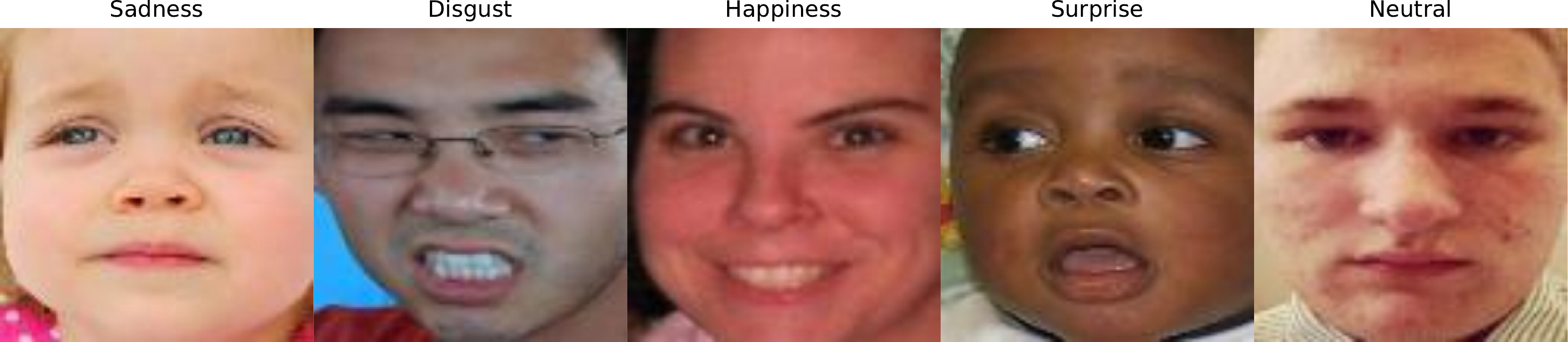}
\caption{Example of face images from the \raf{} dataset. On top of each image, we report the corresponding ground truth expression.}
\label{fig:raf}
\end{figure}

The~\raf{} database comprises 29,672 facial images retrieved from the Internet. Concerning the basic emotion recognition task, the images are manually annotated and distributed in seven different classes: surprise, fear, disgust, happiness, sadness, anger, neutral. Moreover, the dataset comprises twelve classes concerning compound emotions. Regarding the basic emotions only, there are in total 15,339 images divided into two groups: training (12,271 samples) and test (3,068 samples). All the aligned face images have the same resolution of 100x100 pixels. In \autoref{fig:raf}, we report a few examples of facial images from the \raf{} dataset.

\subsection{Oulu-CASIA NIR\&VIS facial expression database (Oulu-CASIA)}\label{subsec:oulu_dataset}
\begin{figure}[!h]
\includegraphics[width=\linewidth]{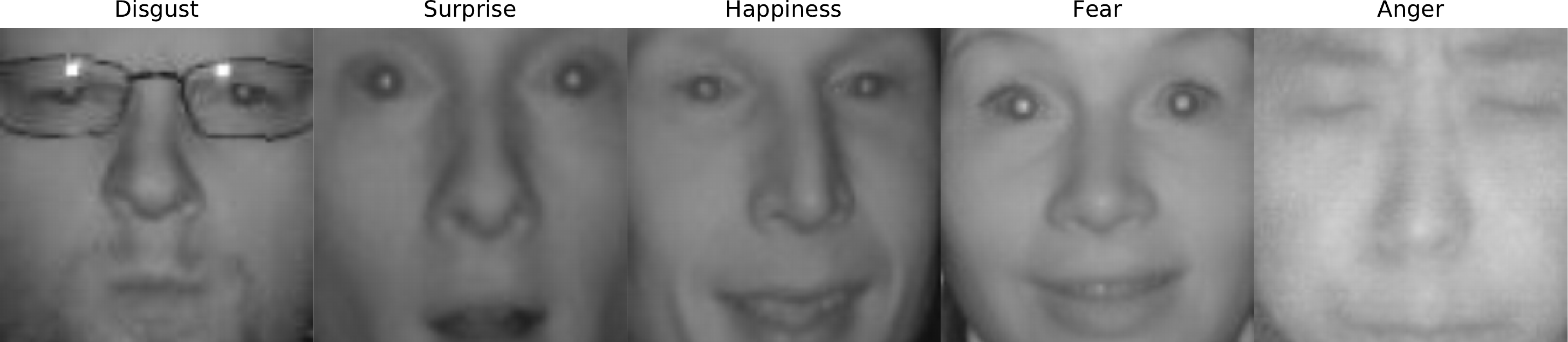}
\caption{Example of face images from the \oulu{} dataset. On top of each image, we report the corresponding ground truth expression.}
\label{fig:oulucasia}
\end{figure}

The~\oulu{} database consists of 2,880 video sequences involving 80 different subjects aging from 23 up to 58 years old. Face expressions are divided into six classes: surprise, happiness, sadness, anger, fear, and disgust. Each video was captured with two imaging systems, i.e., near-infrared (NIR) and visible light (VIS), considering three different illumination conditions, namely normal, weak, and dark. The resolution of the frames extracted from the videos is 320x240 pixels. Specifically, such a resolution concerns the entire frame and not the face only. Indeed, differently from the previous two datasets, i.e., \fer{} and \raf{}, the \oulu{} dataset allows to test in a natively multi-resolution context since the cropped faces have different resolutions. In \autoref{fig:oulucasia}, we report a few examples of facial images from the \oulu{} dataset.

\begin{figure}[!h]
\includegraphics[width=\linewidth]{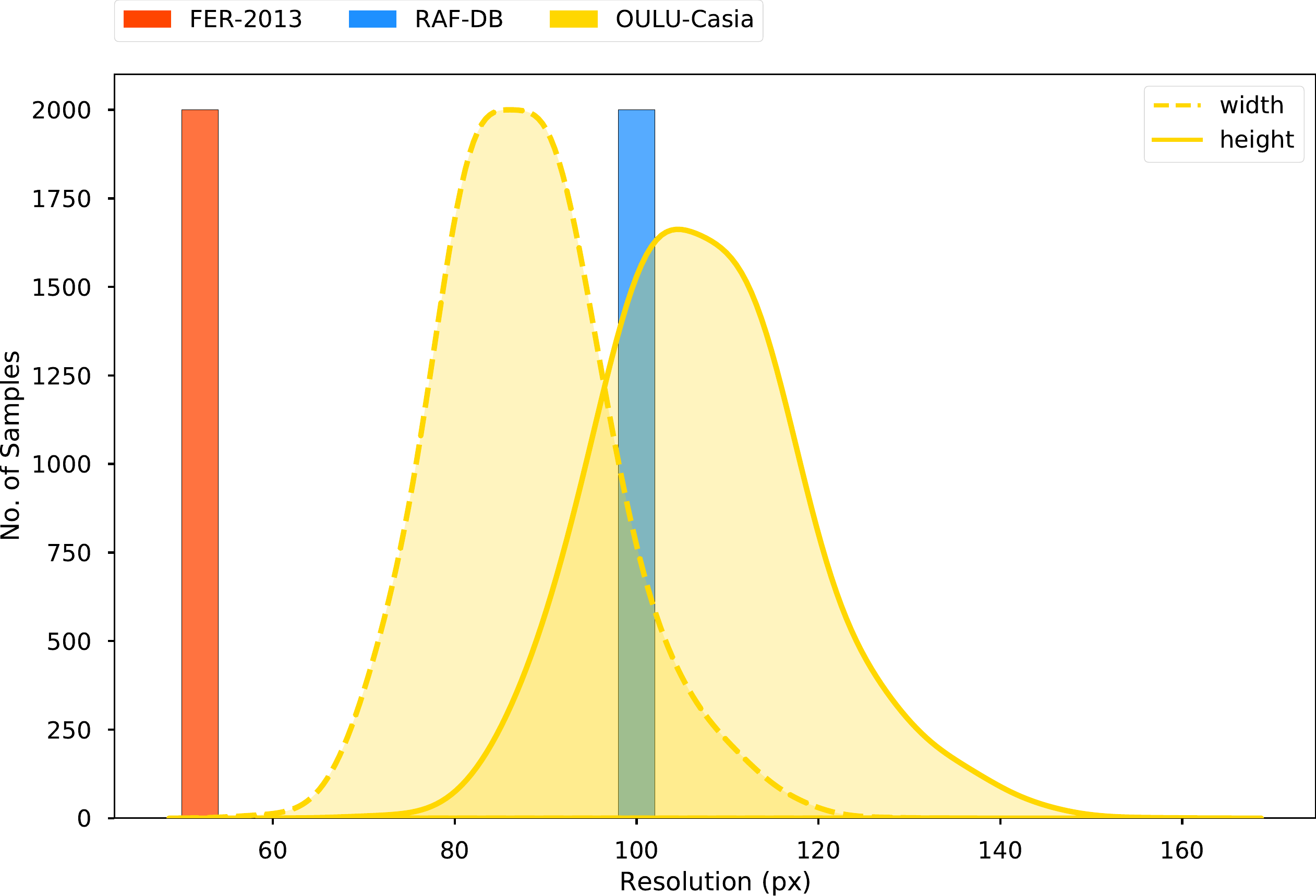}
\caption{Distribution of face resolution concerning the \fer{}, \raf{}, and \oulu{} datasets. The distributions are evaluated considering 2000 images for each dataset.}
\label{fig:dset_res}
\end{figure}

To show the difference among the various datasets that we use to test our training approach, we report the resolution distributions in \autoref{fig:dset_res}. 

As one can notice from \autoref{fig:dset_res}, both \fer{} and \raf{} contain face images at the same resolution, namely, 48x48 and 100x100 pixels, respectively. Instead, \oulu{} is characterized by images with different resolutions, thus allowing us to test our approach on a genuine multi-resolution environment.

%% file: sec/4_training_approach.tex
\section{Training Approach} \label{sec:training_approach}

In this section, we describe in detail our experimental procedure. First, we introduce the architecture of our models and then describe the training approach.

Concerning the learning models, we exploit a ResNet-50~\cite{he2015deep} architecture, equipped with Squeeze-and-Excitation~\cite{hu2018squeeze} blocks, that generates 2048-dimensional deep features vectors used to fulfill the classification task. 

As mentioned in~\autoref{sec:introduction}, our contribution stems from the idea that the FER task might benefit from a multi-resolution training. Thus, we propose the \MODEL{} approach to explicitly leverage images at different resolutions while training a DL model. Specifically, such an approach accounts for a two-step training procedure. First, the DCNN is trained with multi-resolution input data to empower it to generate resolution-robust deep features. Then, a second training step is performed to account for the specific FER task. 

\subsection{Multi-resolution Training}
Concerning the first training phase, we formulate two different flavors for it. First, we leverage the same procedure as in~\cite{massoli2020cross}, and then we use it as a reference to propose a lightweight version of the multi-resolution step to make it less resource-demanding. Specifically, a schematic view of the proposed simplified procedure is shown in~\autoref{fig:sketch}. Compared to the original approach in~\cite{massoli2020cross}, there are two main differences. On the one hand, we do not use the ``Teacher"  supervision signal, and on the other, we use the \affwild{} dataset instead of VGGFace2~\cite{cao2017vggface2}. We want to stress here that we use the \affwild{} dataset for training purposes only because the ground truth for the test is not available yet as the dataset is currently employed in the ABAW 2020 competition~\cite{kollias2020analysing}.
We refer the reader to~\autoref{affwild2dset} for more details on the \affwild{} dataset.

\subsection{Simplified Multi-resolution Training}
We can now focus on the simplified approach to the multi-resolution step proposed in this manuscript (\autoref{fig:sketch}). The two random extractions in the red box are used to decide whether to downsample an image and pick its final resolution. The first one is related to the use of the curriculum learning paradigm~\cite{bengio2009curriculum}, i.e., we define a growing upper bound for the probability of an image to be downsampled. Thus, there is a growing probability for the model to be fed with different resolution images while the training proceeds. Instead, the second extraction is used to pick the resolution at which an input image has to be scaled. However, the face is not modified if the extracted resolution is higher than the initial one. Finally, we want to stress that in the case of the ``simplified" flavor, we use the \affwild{} dataset in the first training step.

\begin{table*}
    \centering
    \begin{tabularx}{\linewidth}{l>{\centering}X>{\centering}X>{\centering}X>{\centering}X>{\centering}X>{\centering}X>{\centering\arraybackslash}X}
    \toprule
    & Neutral &  Anger &  Disgust  &  Fear &  Happiness & Sadness &  Surprise \\
    \cmidrule{2-8}
    Aff-Wild2~\cite{affwild2} & 0.365 & 0.975 & 0.986 & 0.988 & 0.837 & 0.891 & 0.958 \\
    
    FER2013~\cite{fer2013} & 0.827 & 0.861 & 0.985 & 0.857 & 0.749 & 0.832 & 0.890 \\
    
    RAF-DB~\cite{rafdb} & 0.795 & 0.942 & 0.941 & 0.977 & 0.611 & 0.838 & 0.895 \\
    
    Oulu-CASIA~\cite{oulucasia} & - & 0.827 & 0.836 & 0.826 & 0.833 & 0.837 & 0.842 \\
    
    
    \bottomrule
    \end{tabularx}
    \caption{Classes' weight for each dataset. The values reported are referred to the training set only. Note that weights do not need to sum up to one. The lower the weight, the higher the cardinality of the corresponding class.}
    \label{tab:weights}
\end{table*}

\subsection{Full Training Procedure}
As mentioned previously, we use two different formulations for the multi-resolution training phase. Concerning the approach from~\cite{massoli2020cross}, we use the VGGFace2~\cite{cao2017vggface2} dataset, while in the case of the simplified one, we use the \affwild{} dataset. 
As we report in~\autoref{sec:experimental_results}, the two approaches offer comparable results. 

Following the multi-resolution training step, we fine-tuned each model on a given dataset, namely, \fer{}, \raf{}, and \oulu{}. We refer the reader to \autoref{sec:dataset}, for a description of the databases used in this study. Since we aim to show the benefits of a multi-resolution approach to the FER objective, we do not apply any special type of training specifically designed for FER, therefore emphasizing the benefits of using a multi-resolution approach. Hitherto, we show in~\autoref{sec:experimental_results}, that our models reach, and in some cases improve upon, state-of-the-art results concerning carefully designed techniques to fulfill the FER task. Moreover, to better emphasize the impact of the multi-resolution training, we train a base model on the second step of \MODEL{} only and report its performance as a baseline for our study.

\subsection{Training Hyperparameters}

Concerning the first training step on the \affwild{} dataset, we use the Adam~\cite{kingma2014adam} optimizer, set the learning rate to $1.e^{-2}$ and $1.e^{-3}$ for the classifier layer and all the convolutional layers, respectively, and the weight decay to $1.e^{-4}$. We validate the model every 200 train iterations so to avoid overfitting, and we drop the learning rates by a factor of 10 every time the accuracy on the validation set sets on a plateau for more than 2000 training iterations. We set the batch size to 256 and consider images with a resolution down to 16 pixels (on the shortest side). Instead, regarding the first training step on the VGGFace2~\cite{cao2017vggface2} dataset, we utilize the same protocol as in~\cite{massoli2020cross}.

Subsequently, we specialize the models on each of the datasets we use to assess its performance. In all these experiments, we use the Adam~\cite{kingma2014adam} optimizer, set the learning rate to $1.e^{-2}$ and $1.e^{-4}$ for the classifier layer and all the convolutional layers, respectively, and set the weight decay to $5.e^{-4}$. 
The only difference here is that, concerning the base model, i.e., the one trained only with the second phase of \MODEL{}, we set the learning rate to $1.e^{-3}$ and $1.e^{-5}$.
We use a batch size of 256 and validate the model at the end of each training epoch. Finally, we drop the learning rates by a factor of 10 every time the accuracy on the validation set sets on a plateau for more than ten epochs. 

Finally, to allow other researchers to reproduce our results, we will made our code and all the models' checkpoints available on github\footnote{\url{https://github.com/fvmassoli/mafer-multires-facial-expression-recognition}}.

\subsection{Data Preprocessing and Balancing}

Since the datasets we use are not very large, we use data augmentation to preprocess the input data to improve the generalization capabilities of the neural networks. 
To account for the different properties of the various datasets, we use slightly different transformations for each of them. However, the majority of them are common to all of the datasets. We referer the reader to our GitHub repo for the specific code.

Concerning the common transformations, we resize the images to 224 pixels and then we randomly apply the horizontal flip, image gray-scaling, color jitter, and perspective transformations. Finally, we normalize the pixels value.

In almost all the datasets we consider, the number of images for the various expressions is not balanced. For such a reason, we employ a balancing strategy to avoid the models overfitting on the most populated classes. Specifically, we assign to each expression a weight, in the range $[0, 1]$,  evaluated as the inverse of its cardinality. Such weights are then used to balance the cross-entropy loss. In~\autoref{tab:weights}, we report the class cardinality and weights for each of the datasets we employ. 

%% file: sec/5_experimental_results.tex
\section{Experimental Results} \label{sec:experimental_results}

In this section, we report the results of our experimental campaigns. However, we first briefly describe the various metrics we employ. 

\subsection{Metrics}\label{subsec:metrics}
To properly compare our results with other authors, we adopt the most commonly used metrics found in the literature. Concerning the~\fer{}, we report the accuracy on the test set, while on the~\oulu{} dataset, since it is not shipped with a test partition, we use a k-fold protocol to evaluate the average accuracy. Specifically, we set k=10. Finally, concerning the \raf{}, we evaluate the accuracy considering two different protocols: overall accuracy, evaluated on all the classes at once, and average accuracy, evaluated on every single class separately and averaged among them. First, we report the accuracy of the whole test set, then we evaluate the accuracy of each class and then quote the average among them. Finally, we evaluate the Precision@k and the mean Average Precision (mAP) concerning the CBIR context.
The Precision@k is defined as the portion of the retrieved items in the k-set that are relevant and is given by:

\begin{equation}\label{eq:prec_k}
\mathrm{Precision@k} = \frac{(\mathrm{k-Retrieved}\ \cap \  \mathrm{Relevant})}{\mathrm{k-Retrieved}}
\end{equation}

Instead, the mAP 
quantifies how good the model is at answering the query. Mathematically, it is defined as:

\begin{equation}
\mathrm{mAP} = 
\frac{\mathrm{\sum_{q=1}^{Q} AP(q)}}{\mathrm{Q}}
\end{equation}
where $Q$ is the number of queries and AP(q) is the Average Precision for a given query ``q". The AP measures the average relevance scores for a set of items returned as answer to a query. Concerning the set of k-retrieved items, the AP is given by:

\begin{equation}
\mathrm{AP} = 
\frac{\mathrm{1}}{\mathrm{GTP}} 
{\mathrm{\sum_{i=0}^{k} \big(p@i\ \cdot \ rel@i}}\big)
\end{equation}

where GTP is the total number of relevant items, k is the number of retrieved items (that can be equal to the size of the whole dataset), p@i is defined in~\autoref{eq:prec_k}, and rel@i is a relevance function which is 1 if the $i$-th item is relevant and 0 otherwise.



\subsection{Experimental Results}

Concerning the FER task, we report the experimental results on the \fer{}, \oulu{}, and \raf{} datasets in~\autoref{tab:fer2013},~\autoref{tab:oulucasia}, and~\autoref{tab:rafdb}, respectively. In all of the mentioned tables, we report the performance of the models trained with \MODEL{} alongside the best results available in the literature.  

Regarding our models, we quote the results from four different experiments on each dataset. First, we apply the second step only to the ``base'' model to have a baseline to compare with. In our case, we use the Se-ResNet-50 from~\cite{cao2017vggface2} as the base model. Subsequently, we consider three different training configurations. Concerning the first step of \MODEL{}, we first employed the same technique as in~\cite{massoli2020cross}, and we name the model ``CR''. Then, we formulated a lightweight version for the multi-resolution step and named the model as ``CR-Simplified". Finally, to ensure that the gain in the performance of our models is not due to the given dataset used in the first training step, we finetune the ``CR'' model on the \affwild{} dataset and name it ``CR+AffWild2''. Then, all the models are specialized on the given FER dataset during the second step of \MODEL{}.

\input{tables/fer2013}
\input{tables/oulucasia}
\input{tables/rafdb}

 As we can see from~\autoref{tab:fer2013},~\autoref{tab:oulucasia}, and~\autoref{tab:rafdb}, the results reported by the three configurations we tested are compatible among them, thus witnessing that the main source of improvement upon the base model is due to the multi-resolution nature of the training and not to the specific dataset used in such a first training step. 
 We can derive two more fundamental conclusions from our results. First, we acknowledge that our models perform better than the current state-of-the-art when considering natively multi-resolution test scenarios. We can appreciate such a result from~\autoref{tab:oulucasia}. Indeed, among the considered test datasets, \oulu{} is the only one that is shipped with images at very different resolutions, as reported in~\autoref{fig:dset_res}. Especially, we acknowledge that on the \oulu{} dataset, our best model improves upon the current state-of-the-art by $\sim6\%$. Instead, concerning fix-resolution scenarios, i.e., \fer{} and \raf{}, our models deliver utterly comparable performance. Again, it is important to stress that our training approach is completely agnostic of the final task, while the ones we are comparing to were specifically designed for the task at hand. Finally, we can particularly pay attention to the results reported by Mahmoudi et al.~\cite{mahmoudiexp2} in~\autoref{tab:fer2013} and~\autoref{tab:rafdb}. Specifically, from~\autoref{tab:rafdb} we see that the mentioned authors reported a value for the overall accuracy higher than ours concerning images at a resolution of 100x100 pixels. Interestingly, from~\autoref{tab:fer2013}, we see that our models perform better than theirs. We can interpret such results by saying that a multi-resolution approach gives more robust features at low resolutions, as in the case of the \fer{} dataset, than models that are trained at a specific resolution. Perhaps, we can explain such behavior by observing that images at lower resolutions carry less discriminative information than the ones at higher resolutions. Thus, training focused on such input data might lead the model to a worse optimizer compared to when images at high resolution are used, too, as in our case.

\subsection{Models Analysis}

\begin{figure*}[!h]
    \begin{tabular}{cccc}
         \includegraphics[width=0.23\linewidth]{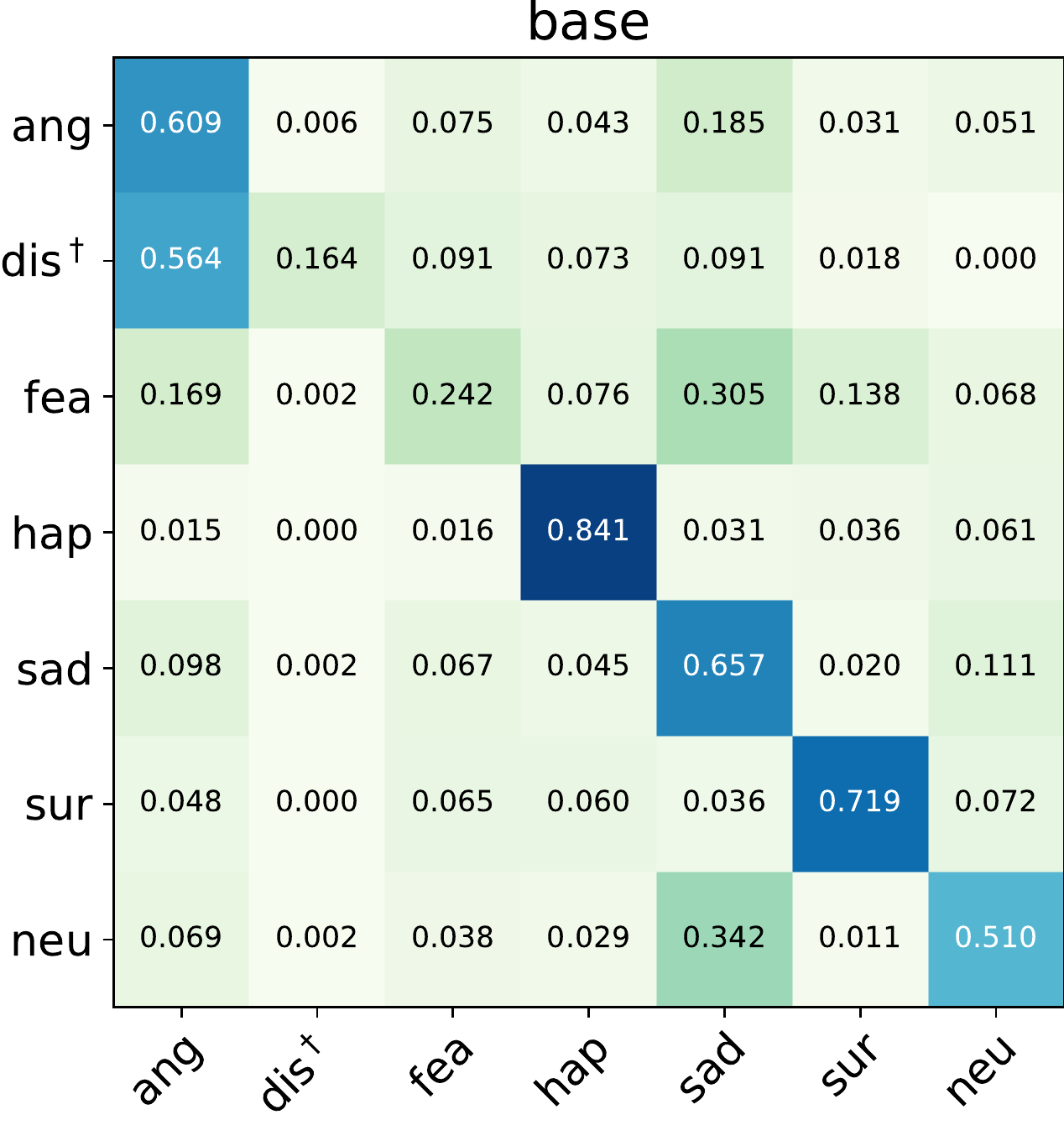}
          & 
          \includegraphics[width=0.23\linewidth]{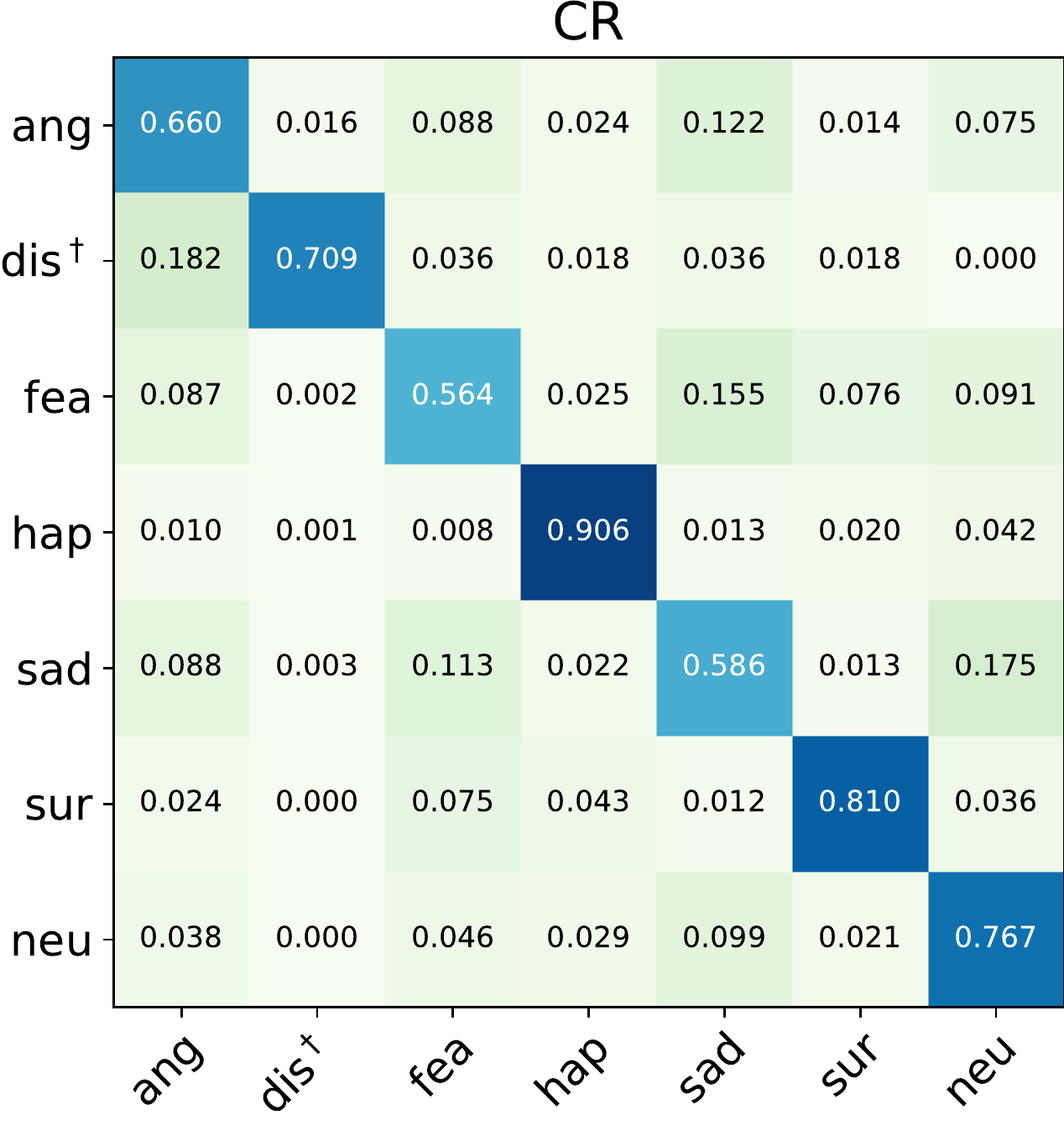}
          & 
          \includegraphics[width=0.23\linewidth]{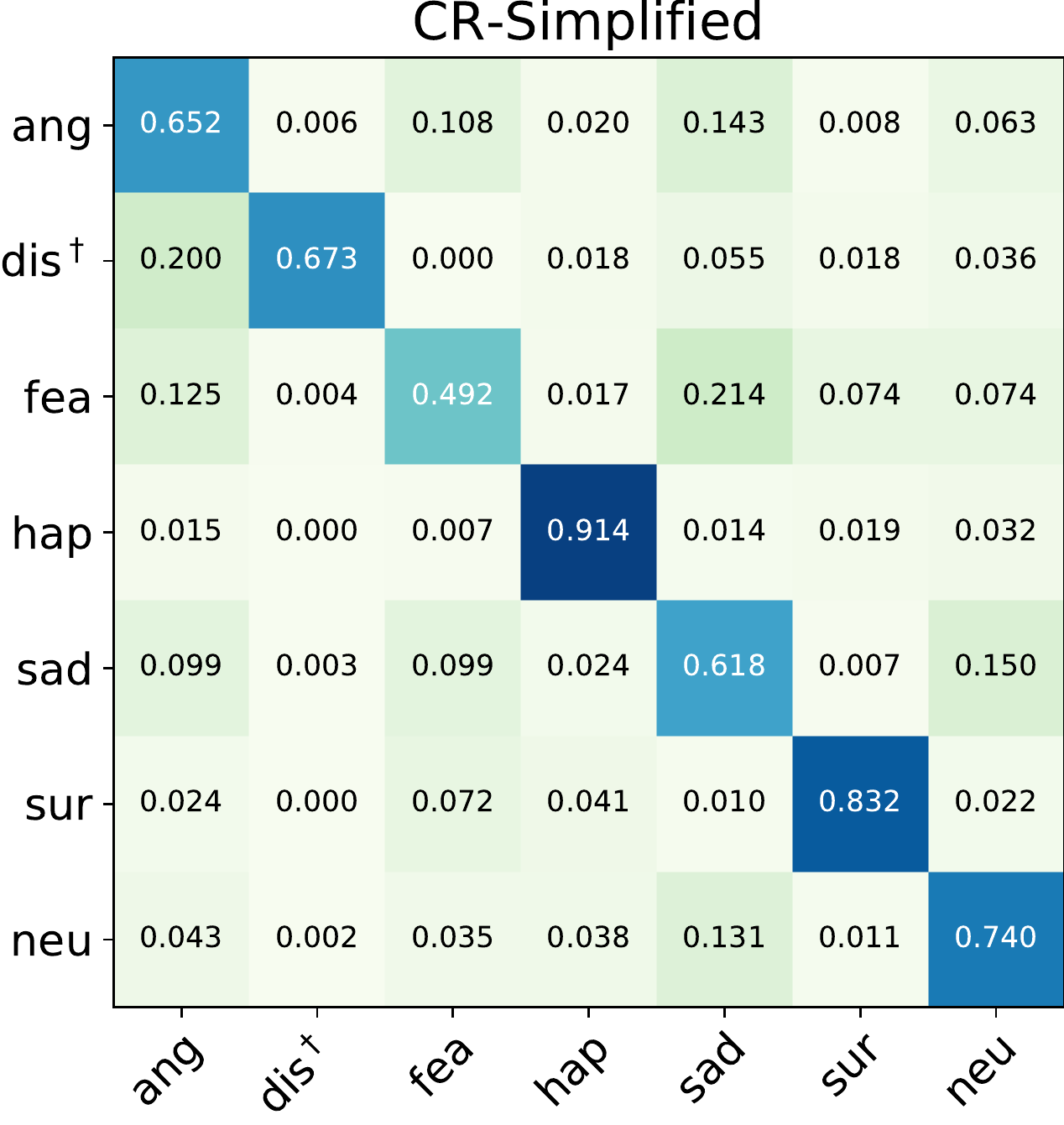}
          &
          \includegraphics[width=0.23\linewidth]{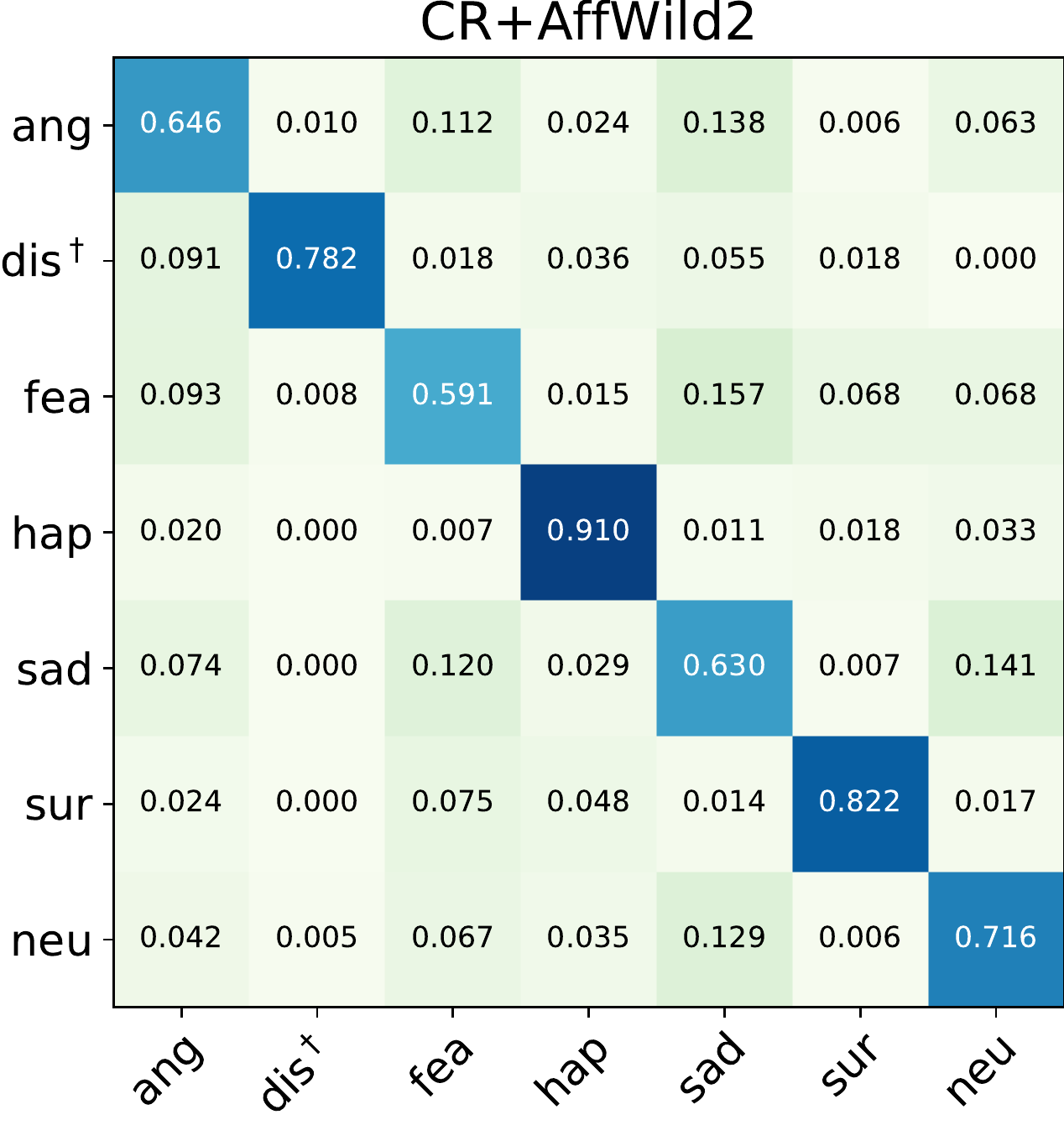}
  \end{tabular}
  \caption{Confusion matrices for the classes' accuracy on the \fer{} dataset. The title of each plot refers to the analyzed model. We emphasize with the symbol `` $\dagger$ " the least populated (\emph{minority}) classes.}
  \label{fig:cm_fer}
\end{figure*}

\begin{figure*}[!h]
    \begin{tabular}{cccc}
        \includegraphics[width=0.23\linewidth]{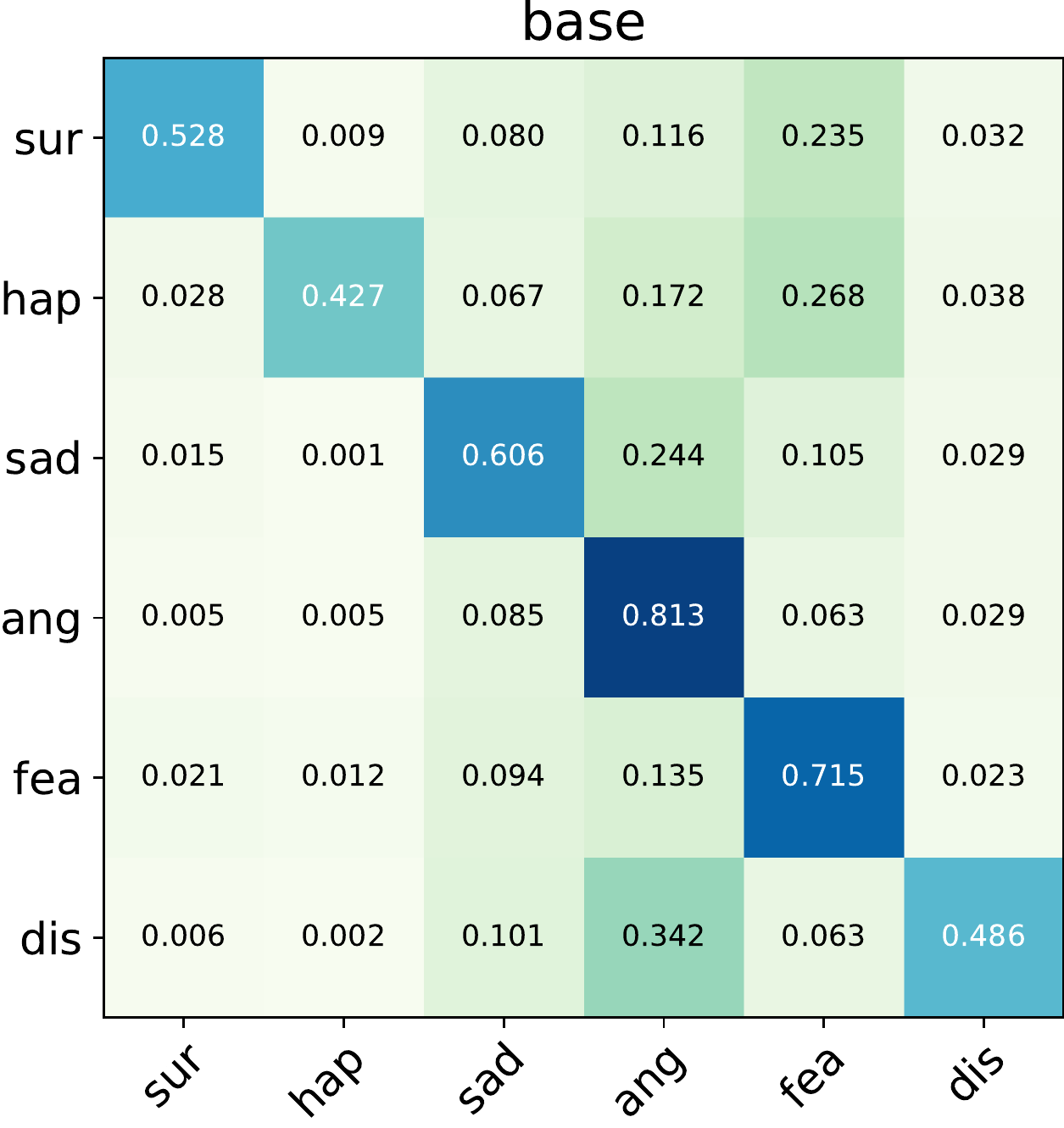}
          & 
          \includegraphics[width=0.23\linewidth]{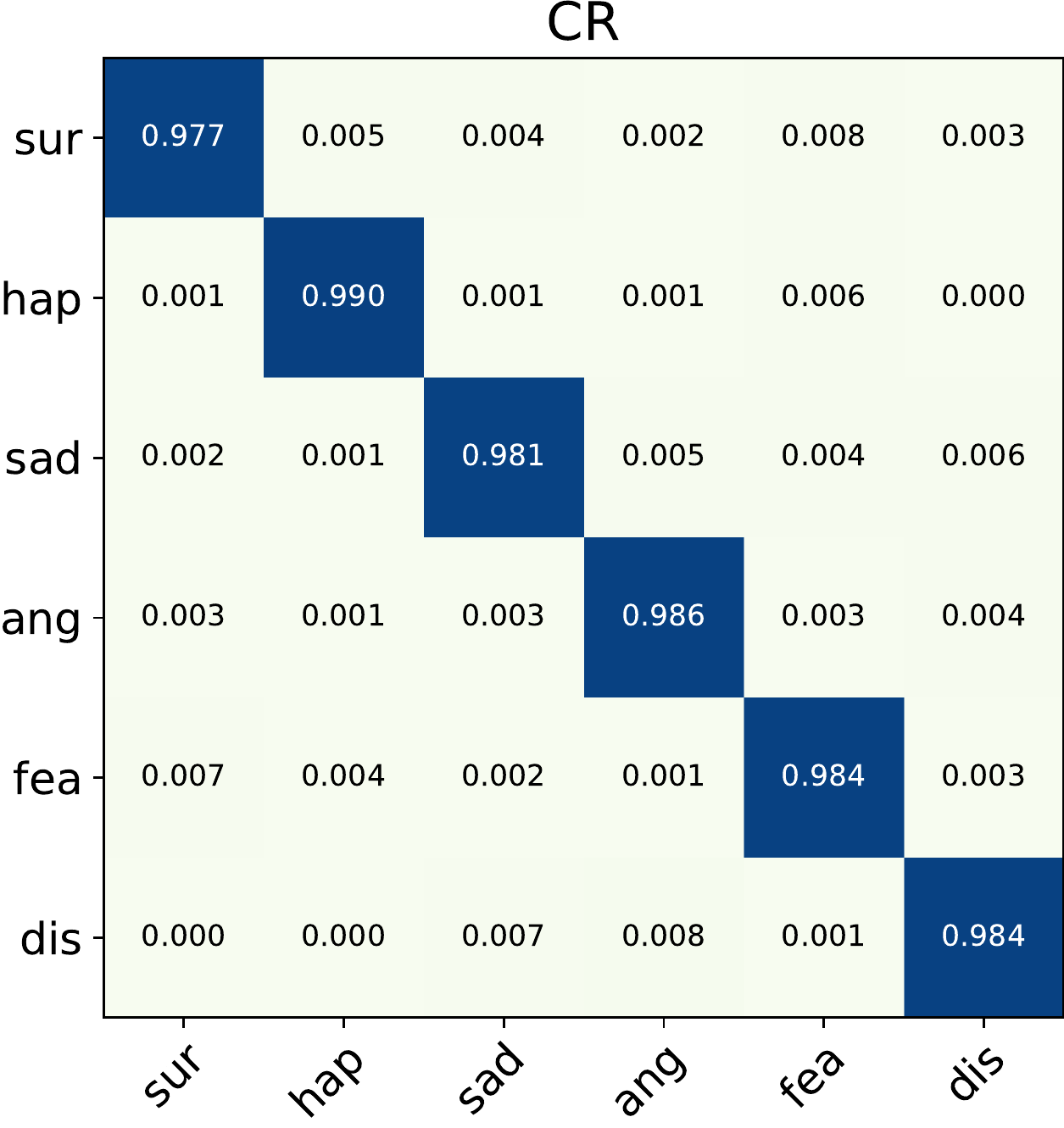}
          & 
          \includegraphics[width=0.23\linewidth]{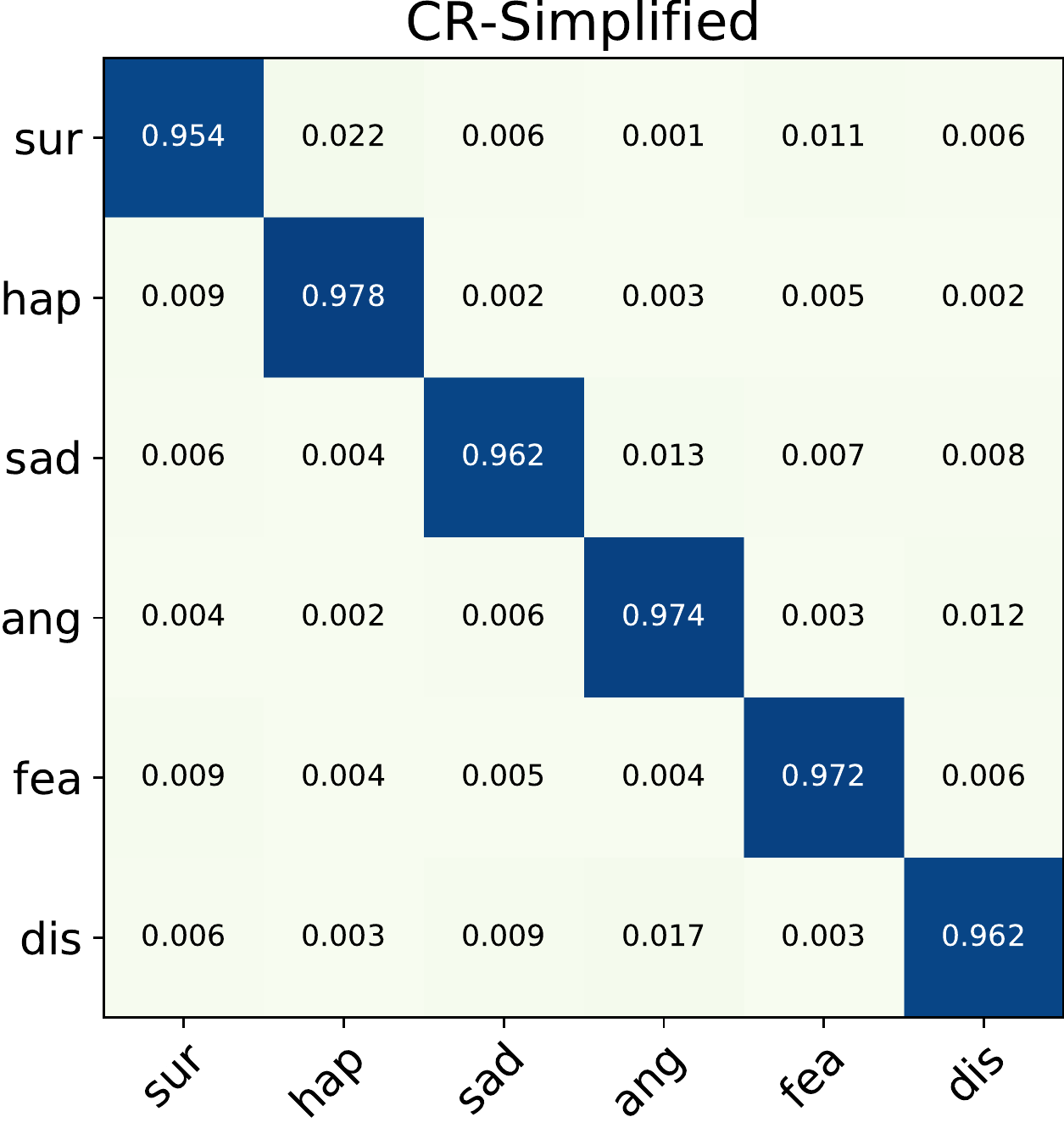}
          &
          \includegraphics[width=0.23\linewidth]{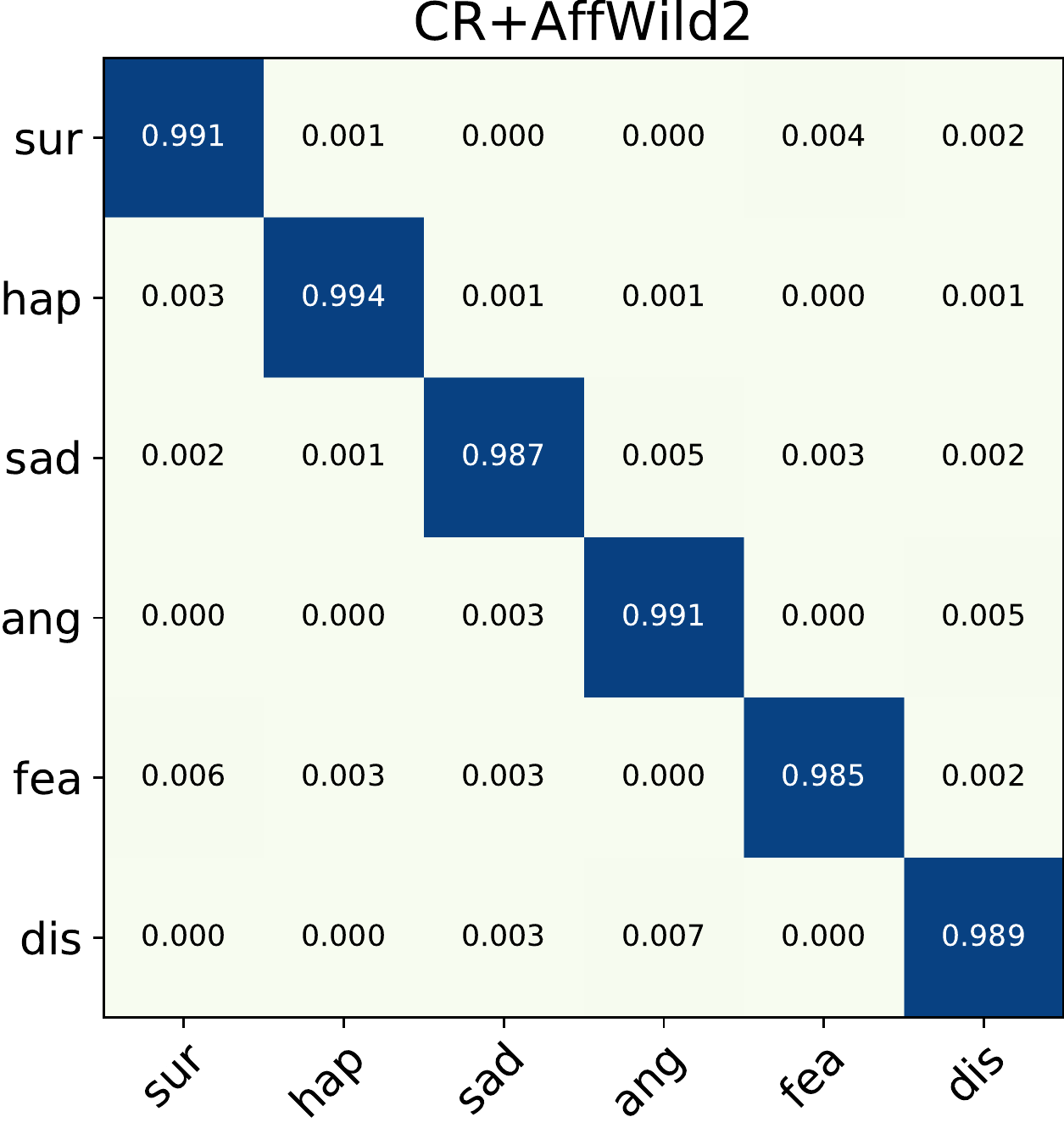}
  \end{tabular}
  \caption{Confusion matrices for the classes' accuracy (10-fold averaged) on the \oulu{} dataset. The title of each plot refers to the analyzed model.}
  \label{fig:cm_oulu}
\end{figure*}

\begin{figure*}[!h]
    \begin{tabular}{cccc}
         \includegraphics[width=0.23\linewidth]{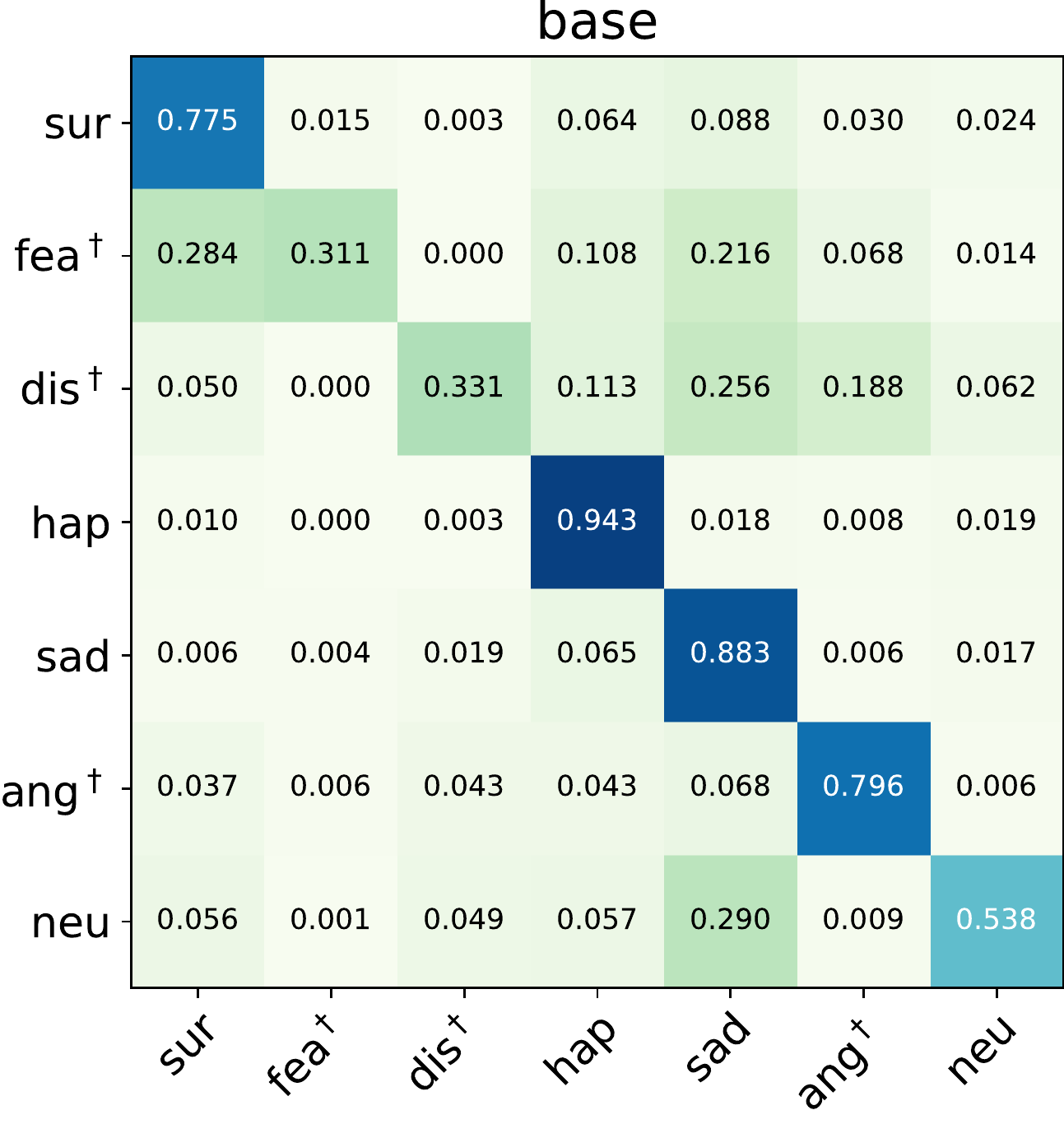}
          & 
          \includegraphics[width=0.23\linewidth]{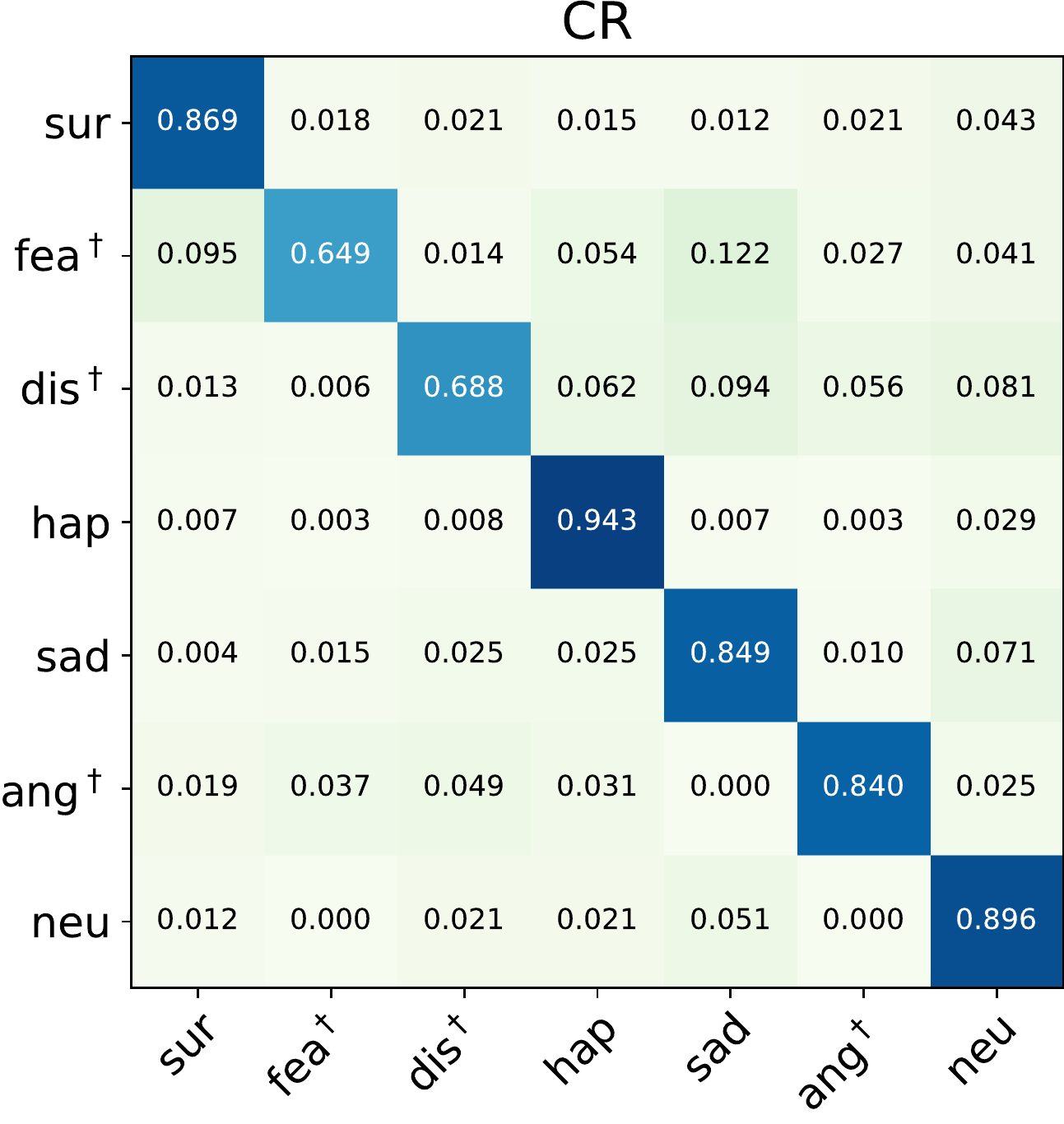}
          & 
          \includegraphics[width=0.23\linewidth]{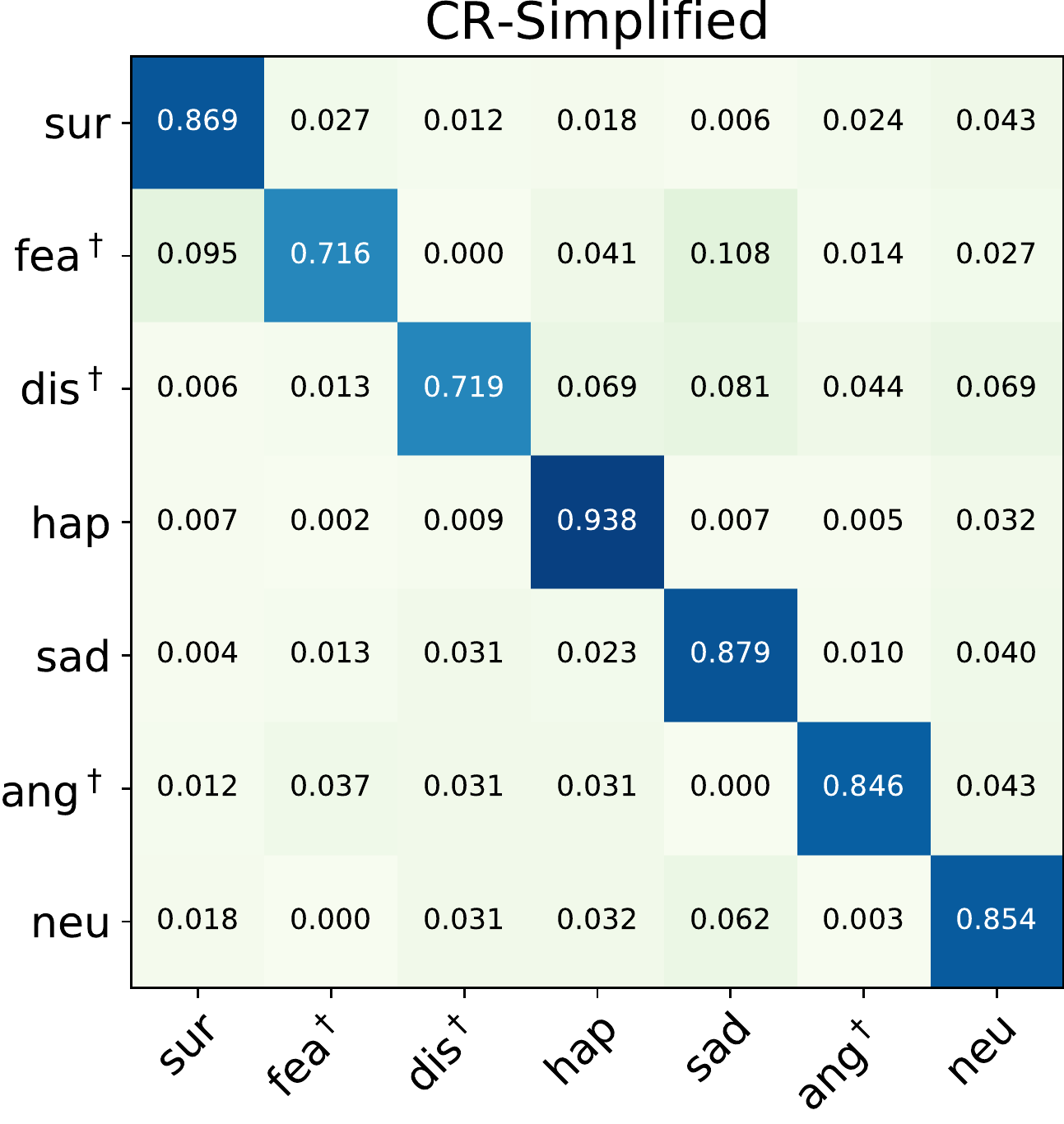}
          &
          \includegraphics[width=0.23\linewidth]{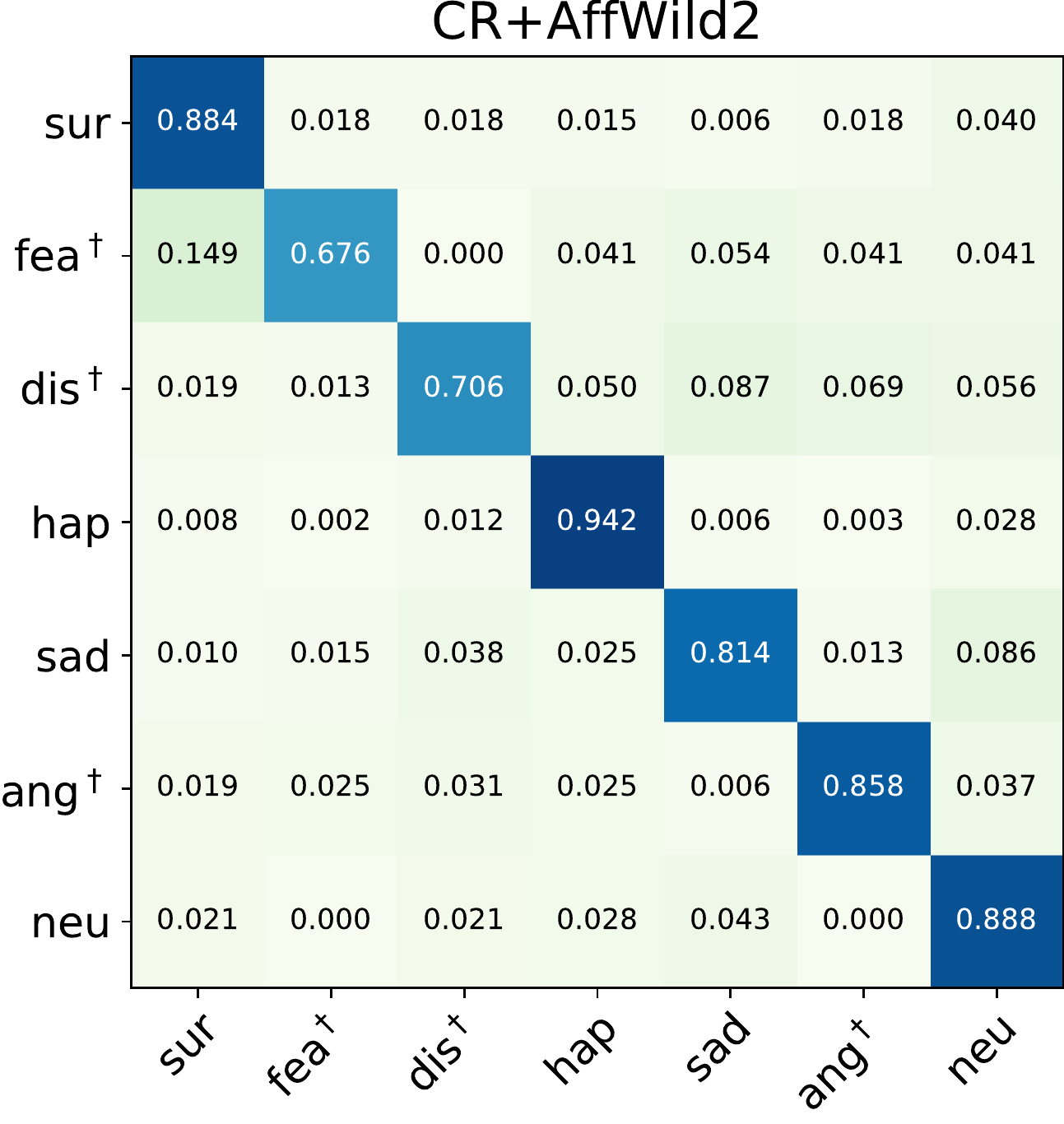}
  \end{tabular}
  \caption{Confusion matrices for the classes' overall accuracy on the \raf{} dataset. The title of each plot refers to the analyzed model. We emphasize with the symbol `` $\dagger$ " the least populated (\emph{minority}) classes.}
  \label{fig:cm_rafdb}
\end{figure*}

In this subsection, we delve deeper into our DCNNs' performances by analyzing their predictions. In~\autoref{fig:cm_fer}, ~\autoref{fig:cm_oulu}, and~\autoref{fig:cm_rafdb}, we report the confusion matrices for all the four models we trained on the \fer{}, \oulu{}, and \raf{} datasets, respectively. By looking at the figures, a common pattern emerges. All the multi-resolution-trained models perform better than the base one on all of the classes. Moreover, we can appreciate that the models behave properly on all classes, meaning that we properly balanced them during the training. The only exception is accuracies on the ``fear'' and ``disgust'' classes of the \raf{} dataset. Specifically, concerning the two mentioned expressions, most mistakes are due to their misclassification as ``sadness''.

\begin{figure*}
\begin{tabular}{cccc}
\includegraphics[width=0.23\linewidth]{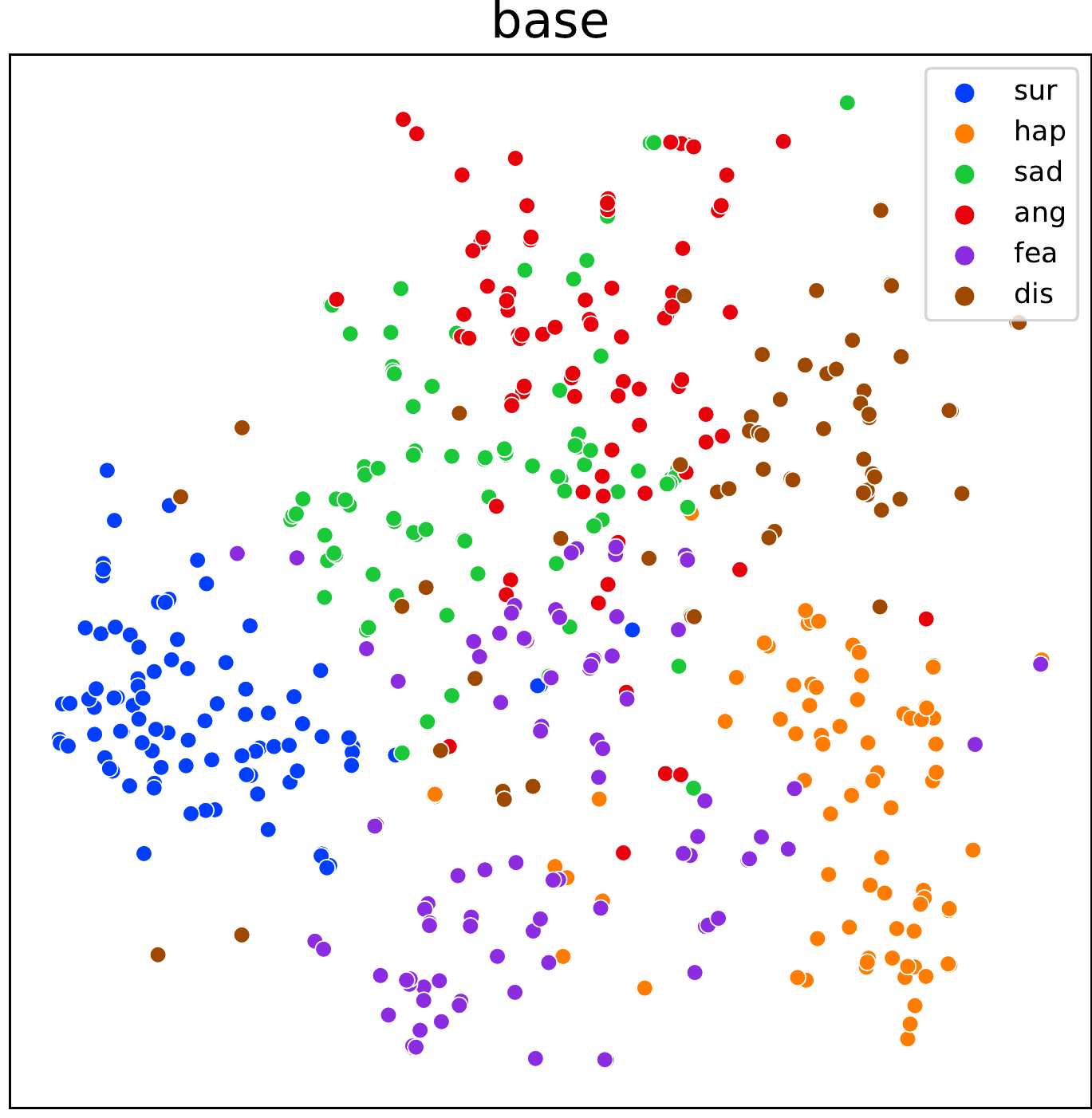} & 
\includegraphics[width=0.23\linewidth]{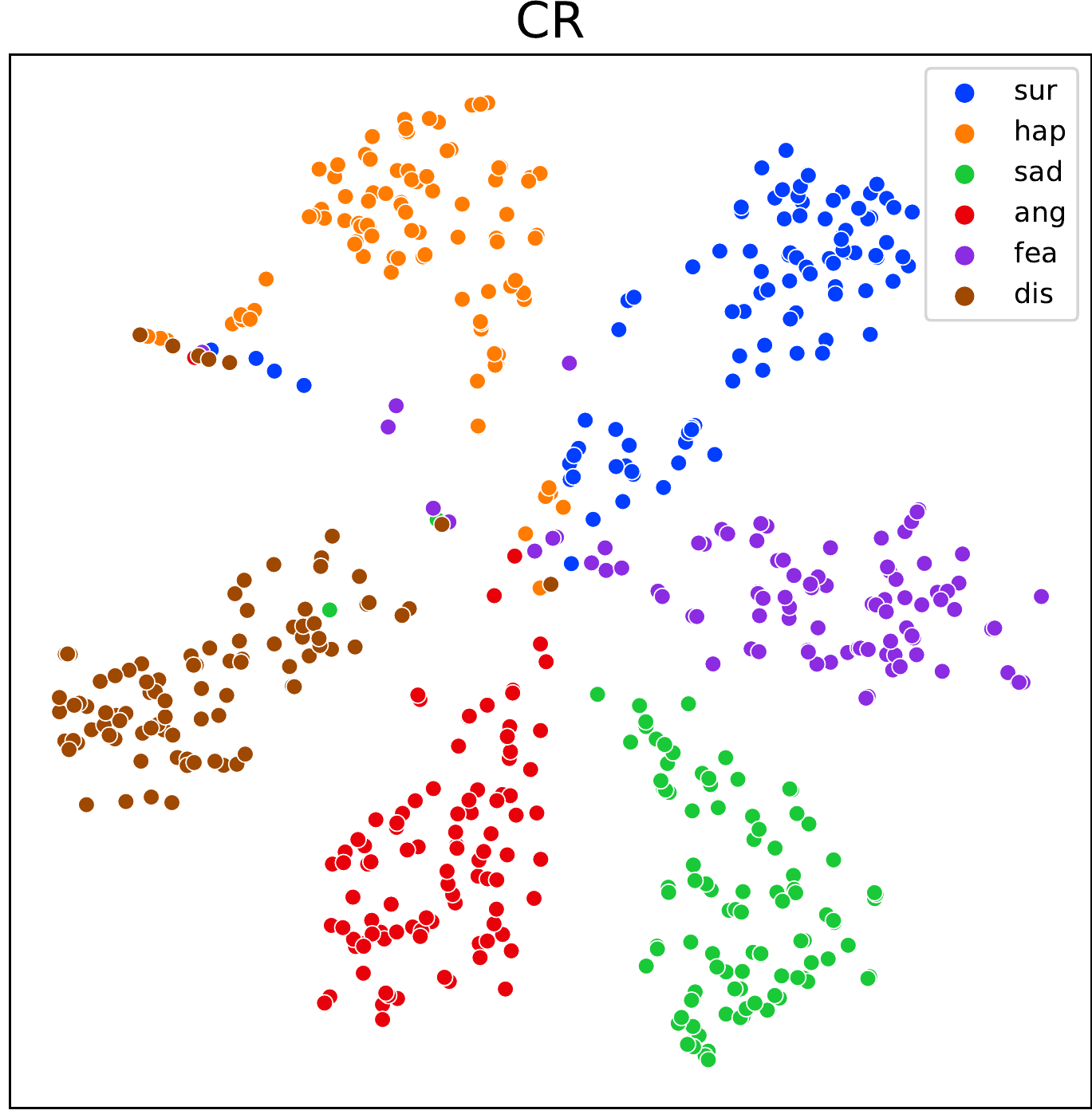}& 
\includegraphics[width=0.23\linewidth]{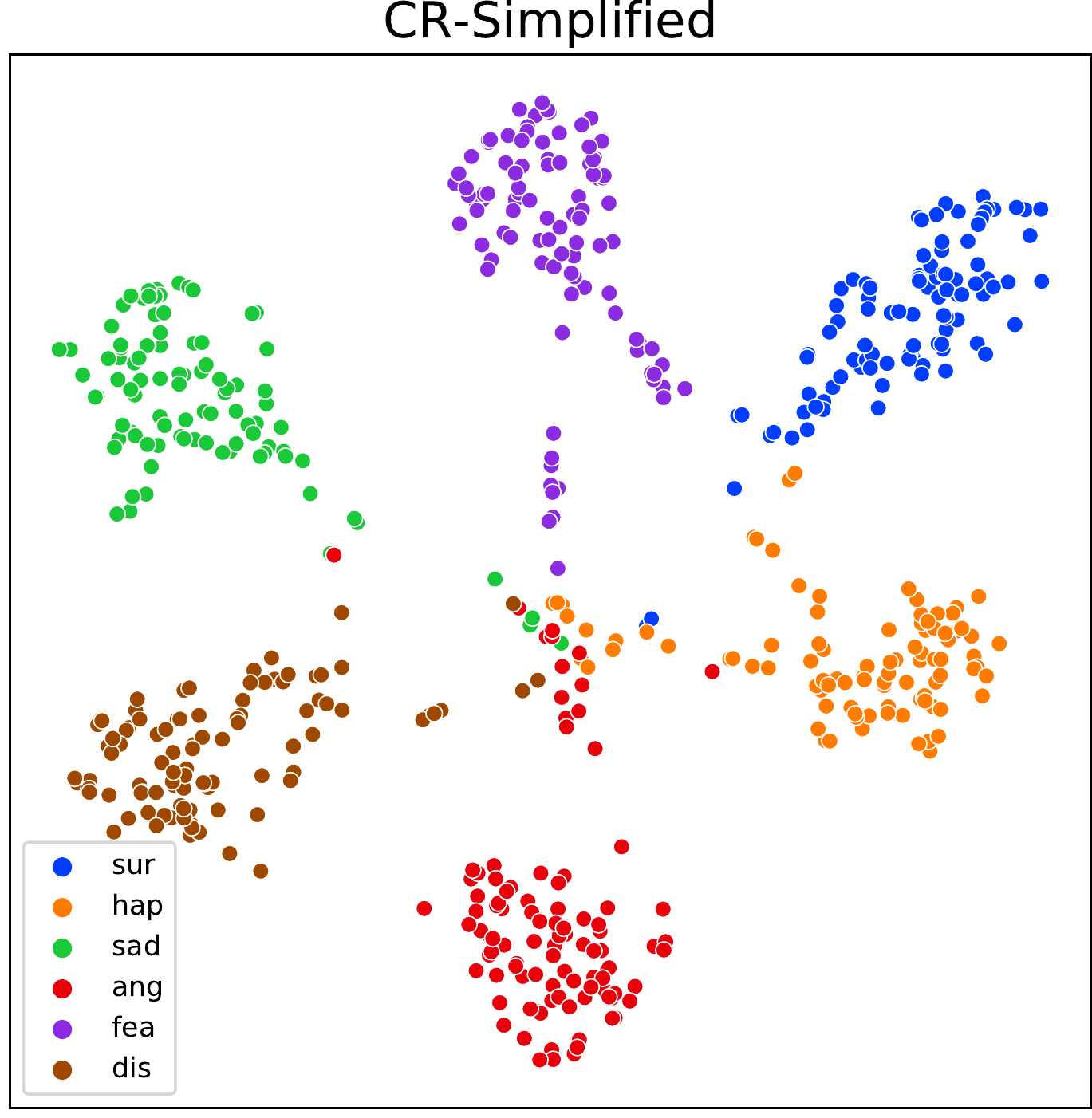}& 
\includegraphics[width=0.23\linewidth]{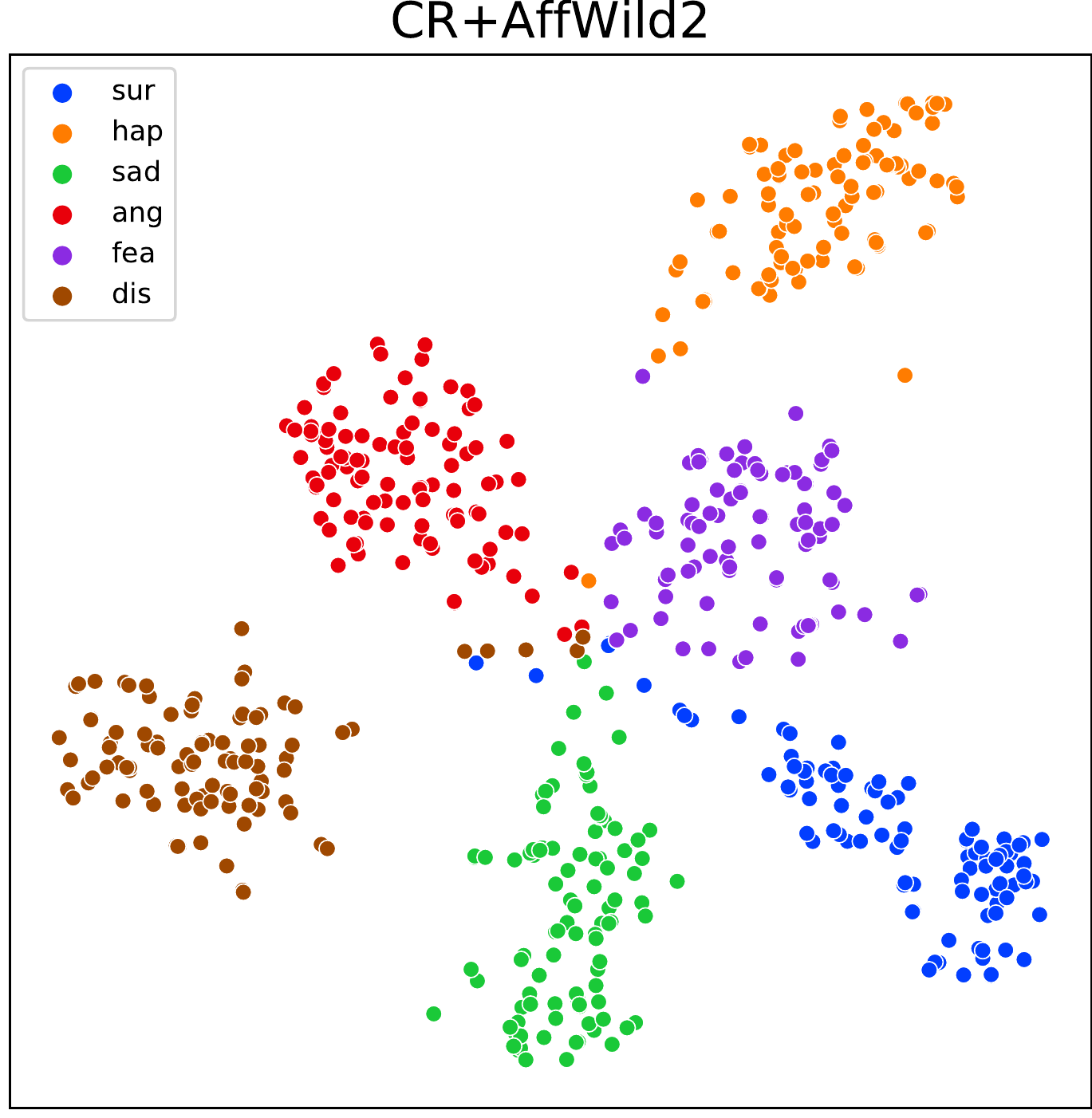}
\end{tabular}
\caption{t-SNE~\cite{maaten2008visualizing} comparison of features generated by models trained on the \oulu{} dataset. The title of each plot refers to the analyzed model. We use 100 images for each class.}
\label{fig:tsne_oulu}
\end{figure*}

Subsequently, we exploit the t-SNE~\cite{maaten2008visualizing} technique to visually inspect and compare the deep representations generated by the base model and the ones trained with \MODEL{}. The results are reported in~\autoref{fig:tsne_oulu} concerning the \oulu{} dataset.

As we can see from~\autoref{fig:tsne_oulu}, the base model cannot separate well the various classes while the other models do a pretty good job. Such a result is of fundamental relevance since we generate the representations using images from the \oulu{} dataset that natively offers a multi-resolution test environment being its images characterized by different resolutions~\autoref{fig:dset_res}. Moreover, by comparing the multi-resolution models, we can see that ``CR-Simplified" performs comparably to the others, thus sustaining our proposal of a simplified version of the multi-resolution training step. 

\subsection{Content-Based Image Retrieval}

Finally, we conclude the section on the experimental results by testing our models against the CBIR task~\cite{donahue2014decaf,class_retrieval,cnn_1,cnn_2,donahue2014decaf} on the \oulu{} dataset. We choose such that dataset since, as reported in~\autoref{subsec:oulu_dataset}, it contains images at different resolutions. As reported in~\autoref{subsec:metrics}, we evaluate the Precision@k and mAP. Concerning the first one, we report results considering $\mathrm{k}\in\{1, 5, 10, 50, 100\}$. To our knowledge, we are the first to perform such a test. Thus we cannot compare to results from other authors. However, we assess the benefits of the multi-resolution training by comparing the ``base" model results. Concerning the CBIR protocol, we apply the following procedure. We randomly select 20 correctly classified images for each expression class for each model to use as queries. Then, the remaining faces are used to train a k-NN classifier. Concerning the k-NN, we use the implementation available from the python package scikit-learn, use the euclidean distance as a similarity measure, and set k=3. 

Concerning the Precision@k and mAP, we report the results in~\autoref{tab:cbir_oulu_prec}.



\input{tables/cbir_oulu_prec}

As we can from~\autoref{tab:cbir_oulu_prec}, the performances of the models trained with \MODEL{} are superior to the base model, thus confirming the effectiveness of a multi-resolution approach. Moreover, we can appreciate that the model learned with the ``simplified" version of the training reports comparable performance to the other ``CR-" models, thus confirming our hypothesis that, in the case of FER, it is possible to simplify the approach proposed in~\cite{massoli2020cross}. Especially, we observe that the performances of the ``CR-Simplified" and ``CR+AffWild2" agree within the errors.

%% file: tables/fer2013.tex
\begin{table}[!h]
    \centering
    \begin{tabularx}{\linewidth}{l>{\centering\arraybackslash}X}
    \toprule
    \textbf{Authors} & \textbf{Accuracy (\%)} \\
    \cmidrule{1-2} 
    Tang et al.~\cite{tang2013deep} & 71.20 \\
    Mahmoudi et al.~\cite{mahmoudiexp2} & 71.35 \\
    Yu et al.~\cite{yu2015image}$^{\star \dagger}$ & (\emph{72.00}) \\
    Kim et al.~\cite{kimfer}$^\dagger$ & (\emph{72.72}) \\
    Connie et al.~\cite{conniefer}& 73.40 \\
    Luan et al.~\cite{fer2013result} & \textbf{74.14} \\
    Luan et al.~\cite{fer2013result}$^\dagger$ & (\emph{76.82}) \\
    
    
    
    base & 60.82 \\
    \\
    MAFAR & \\
    \cmidrule{1-1} 
    CR &  73.06 \\
    CR-Simiplified  & 72.33 \\
    CR+AffWild2  & 73.45 \\ 
    
    \bottomrule
    \multicolumn{2}{@{}l}{$^\star$Results from~\cite{georgescufer}} \\
    \multicolumn{2}{@{}l}{$^\dagger$Should not be directly compared with others since the results comes from} \\
    \multicolumn{2}{@{}l}{ensembles of models.} \\
    
    \end{tabularx}
  \caption{Results on the FER2013~\cite{fer2013} dataset. We emphasize in bold the performance of the best model.} 
   \label{tab:fer2013}
\end{table}

%% file: tables/oulucasia.tex
\begin{table}[!h]
    \centering
    \begin{tabularx}{\linewidth}{l>{\centering\arraybackslash}X}
    \toprule
    \textbf{Authors} & \textbf{Accuracy (\%)} \\
    & \textbf{10-fold Averaged} \\
    \cmidrule{1-2} 
    Zhao et al.~\cite{zhao2016peakoulu} & 84.59 \\
    Ding et al.~\cite{ding2oulu} & 87.71 \\
    Zhang et al.~\cite{zhang2017facialoulu} & 86.25 \\
    Zhang et al.~\cite{zhang2020oulu} & 86.00 \\
    Ming et al.~\cite{Oulu-CASIAresult} & 89.60 \\
    Yu et al.~\cite{yuoulu} & 90.40 \\
    Yang et al.~\cite{yang2017facialoulu} & 92.89 \\
    base & 59.84  $\pm$ 1.29 \\
    \\
    MAFAR & \\
    \cmidrule{1-1} 
    CR & \textbf{98.40 $\pm$ .11} \\
    CR-Simplified  & \textbf{96.72 $\pm$ .24} \\
    CR+AffWild2  & \underline{\textbf{98.95 $\pm$ .15}} \\
    \bottomrule
    \end{tabularx}
    \caption{Results on the Oulu-CASIA~\cite{oulucasia} dataset. We emphasize in bold the performance of the best model. If we improve upon the current state-of-the-art with more than one model, we underline the highest performance.}
    \label{tab:oulucasia}
\end{table}

%% file: tables/rafdb.tex
\begin{table}[!h]
\centering
\begin{tabularx}{\linewidth}{l>{\centering}X>{\centering}X>{\centering\arraybackslash}X}
\toprule
\textbf{Authors} & \multicolumn{3}{c}{\textbf{Accuracy (\%)}} \\ \cmidrule{2-4} 
 & \textbf{Overall} &  & \textbf{Average}\\ 
\cmidrule{1-4} 
Vo et al.~\cite{raf-dbresultvo} & 88.98  & & 80.78 \\
Farzaneh et al.~\cite{raf-dbresultfarz} & 87.78  & & 80.44 \\
Farzaneh et al.~\cite{farzanehraf-db} & 86.90  & & 79.71 \\
Florea et al.~\cite{florearaf-fb} & 84.50  & & 76.50 \\
Mahmoudi et al.~\cite{mahmoudiexp2} & \textbf{93.21}  & & - \\
Wang et al.~\cite{wang2020regionraf} & 86.90  & & - \\
base &  77.09 & & 65.39 $\pm$ 9.83 \\
\\
    MAFAR & \\
    \cmidrule{1-1} 
    CR & 88.43 && \textbf{81.90 $\pm$ 4.12} \\
    CR-Simplified  & 88.14 & & \underline{\textbf{83.16 $\pm$ 2.92}} \\
    CR+AffWild2  & 88.07 & & \textbf{82.40 $\pm$ 3.74}  \\
\bottomrule
\end{tabularx}
\caption{Results on the RAF-DB~\cite{rafdb} dataset. We emphasize in bold the performance of the best model. If we improve upon the current state-of-the-art with more than one model, we underline the highest performance.}
\label{tab:rafdb}
\end{table}

%% file: tables/cbir_oulu_prec.tex
\begin{table*}[!h]
\centering
\begin{tabularx}{\linewidth}{lc>{\centering}X>{\centering}X>{\centering}X>{\centering}X>{\centering}X>{\centering}X>{\centering\arraybackslash}X}
\toprule
\textbf{Model} & & \multicolumn{5}{c}{\textbf{Precision@k (\%)}} & & \textbf{mAP} \\
\cmidrule{3-7}
 & & 1 & 5 & 10 & 50 & 100 & &\\
\hline

base          & & 83.3 $\pm$ 3.4 & 67.0 $\pm$ 2.9 & 60.2 $\pm$ 2.5 & 49.0 $\pm$ 1.8 & 42.2 $\pm$ 1.4 & &  35.4 $\pm$ 1.1 \\ 

\\
MAFAR & & & & & & & & \\ \cmidrule{1-1} 

CR            & & 95.8 $\pm$ 1.8 & 92.2 $\pm$ 2.0 & 89.8 $\pm$ 2.2 & 84.4 $\pm$ 2.5 & 80.3 $\pm$ 2.4 & & 70.8 $\pm$ 1.5 \\ 

CR-Simplified & & \textbf{100} $\pm$ 0.0 & \textbf{98.5 $\pm$ 0.8} & 96.9 $\pm$ 1.0 & 91.5 $\pm$ 1.8 & 88.0 $\pm$ 2.0 & & 78.6 $\pm$ 1.5 \\ 

CR+AffWild2   & & 99.2 $\pm$ 0.8 & \textbf{98.5 $\pm$ 0.7} & \textbf{97.8 $\pm$ 0.8} & \textbf{93.6 $\pm$ 1.3} & \textbf{89.5 $\pm$ 1.5} & & \textbf{79.6 $\pm$ 1.7} \\ 

\bottomrule
\end{tabularx}
\caption{Precision@k and mAP for the \oulu{} dataset. Concerning the Precision@k, we report results for $\mathrm{k}\in\{1, 5, 10, 50, 100\}$. Each value represents the average over 120 queries (20 for each class). We emphasize in bold the performance of the best model.}
\label{tab:cbir_oulu_prec}
\end{table*}

%% file: sec/6_conclusions.tex
\section{Conclusion} \label{sec:conclusion}

Within the deep learning community, the automatic analysis of human facial expressions is an active research area, primarily due to its relevance to several applications as human-computer interactions, customer marketing, and health monitoring, among others. In such a context, we conjecture that deep learning models tasked with FER might benefit from a multi-resolution training approach. To prove that it is indeed the case, we conducted an extensive experimental campaign on publicly available datasets, namely, FER2013, RAF-DB, and Oulu-CASIA. Thus, we propose a two-step training procedure, named \MODEL{}, which directly leverages a multi-resolution learning technique. By adopting \MODEL{}, our models improve upon the current state-of-the-art in multi-resolution scenarios and obtain utterly comparable performance in fix-resolution contexts. Notably, we acknowledge the $\sim6\%$ improvement of the accuracy on the Oulu-CASIA dataset. To better emphasize the benefits of the multi-resolution training step, we trained our models with four different configurations, namely, ``base", ``CR", ``CR-Simplified", and ``CR+AffWild2". By looking at the obtained results, we can conclude that the multi-resolution training empowers deep learning models to generate predictions for FER that are robust against a wide range of resolutions. Finally, we analyze our models to assess their performance better. As a first remark, we observe that the behavior of our models is well balanced among the various classes. Then, we visually inspect the quality of the deep features, i.e., how well they can be separated when belonging to different classes, by exploiting the t-SNE technique. From such an analysis, one can notice that the features generated by the model trained with \MODEL{} are much better separated than the ones generated by the ``base" model. We can interpret such a result by asserting that \MODEL{} empowers deep learning models to generate discriminative features robust against variations in the resolutions of the input images. Finally, to numerically sustain the higher discriminability of deep representations generated by models trained with \MODEL{} compared to the base model,  we quote the Precision@k, Recall@k, and mAP in the context of CBIR. To perform such tests, we use the OULU-Casia dataset since it contains images at different resolutions. The experimental results confirm, also in this case, that the image resolution plays a relevant role in the context of FER and that learning models benefit from a  multi-resolution approach.

%% file: main.bbl
\begin{thebibliography}{10}
\providecommand{\url}[1]{\texttt{#1}}
\providecommand{\urlprefix}{URL }
\providecommand{\doi}[1]{https://doi.org/#1}

\bibitem{alrubaish2020effects}
Alrubaish, H.A., Zagrouba, R.: The effects of facial expressions on face
  biometric system’s reliability. Information  \textbf{11}(10), ~485 (2020)

\bibitem{cnn_1}
Babenko, A., Slesarev, A., Chigorin, A., Lempitsky, V.: Neural codes for image
  retrieval. In: European conference on computer vision. pp. 584--599. Springer
  (2014)

\bibitem{bartlett2003real}
Bartlett, M.S., Littlewort, G., Fasel, I., Movellan, J.R.: Real time face
  detection and facial expression recognition: development and applications to
  human computer interaction. In: 2003 CVPR workshop. vol.~5, pp. 53--53. IEEE
  (2003)

\bibitem{bengio2009curriculum}
Bengio, Y., Louradour, J., Collobert, R., Weston, J.: Curriculum learning. In:
  Proceedings of the 26th annual international conference on machine learning.
  pp. 41--48. ACM (2009)

\bibitem{cao2017vggface2}
Cao, Q., Shen, L., Xie, W., Parkhi, O.M., Zisserman, A.: Vggface2: A dataset
  for recognising faces across pose and age. corr abs/1710.08092 (2017).
  arXiv:1710.08092  (2017)

\bibitem{card2018psychology}
Card, S.K.: The psychology of human-computer interaction. Crc Press (2018)

\bibitem{chen2014facial}
Chen, J., Chen, Z., Chi, Z., Fu, H., et~al.: Facial expression recognition
  based on facial components detection and hog features. In: International
  workshops on electrical and computer engineering subfields. pp. 884--888
  (2014)

\bibitem{conniefer}
Connie, T., Al-Shabi, M., Cheah, W.P., Goh, M.: Facial expression recognition
  using a hybrid cnn--sift aggregator. In: International Workshop on
  Multi-disciplinary Trends in Artificial Intelligence. pp. 139--149. Springer
  (2017)

\bibitem{cowie2003describing}
Cowie, R., Cornelius, R.R.: Describing the emotional states that are expressed
  in speech. Speech communication  \textbf{40}(1-2),  5--32 (2003)

\bibitem{dalgleish2000handbook}
Dalgleish, T., Power, M.: Handbook of cognition and emotion. John Wiley \& Sons
  (2000)

\bibitem{ding2oulu}
Ding, H., Zhou, S.K., Chellappa, R.: Facenet2expnet: Regularizing a deep face
  recognition net for expression recognition. In: 2017 12th IEEE International
  Conference on Automatic Face \& Gesture Recognition (FG 2017). pp. 118--126.
  IEEE (2017)

\bibitem{donahue2014decaf}
Donahue, J., Jia, Y., Vinyals, O., Hoffman, J., Zhang, N., Tzeng, E., Darrell,
  T.: Decaf: A deep convolutional activation feature for generic visual
  recognition. In: International conference on machine learning. pp. 647--655
  (2014)

\bibitem{ekman1992argument}
Ekman, P.: An argument for basic emotions. Cognition \& emotion
  \textbf{6}(3-4),  169--200 (1992)

\bibitem{ekman1999basic}
Ekman, P.: Basic emotions. Handbook of cognition and emotion
  \textbf{98}(45-60), ~16 (1999)

\bibitem{farzanehraf-db}
Farzaneh, A.H., Qi, X.: Discriminant distribution-agnostic loss for facial
  expression recognition in the wild. In: Proceedings of the IEEE/CVF CVPR
  Workshops. pp. 406--407 (2020)

\bibitem{raf-dbresultfarz}
Farzaneh, A.H., Qi, X.: Facial expression recognition in the wild via deep
  attentive center loss. In: Proceedings of the IEEE/CVF Winter Conference on
  Applications of Computer Vision. pp. 2402--2411 (2021)

\bibitem{fei2020deep}
Fei, Z., Yang, E., Li, D.D.U., Butler, S., Ijomah, W., Li, X., Zhou, H.: Deep
  convolution network based emotion analysis towards mental health care.
  Neurocomputing  \textbf{388},  212--227 (2020)

\bibitem{florearaf-fb}
Florea, C., Florea, L., Badea, M.S., Vertan, C., Racoviteanu, A.: Annealed
  label transfer for face expression recognition. In: BMVC. p.~104 (2019)

\bibitem{generosi2018deep}
Generosi, A., Ceccacci, S., Mengoni, M.: A deep learning-based system to track
  and analyze customer behavior in retail store. In: 2018 IEEE 8th
  International Conference on Consumer Electronics-Berlin (ICCE-Berlin).
  pp.~1--6. IEEE (2018)

\bibitem{georgescufer}
Georgescu, M.I., Ionescu, R.T., Popescu, M.: Local learning with deep and
  handcrafted features for facial expression recognition. IEEE Access
  \textbf{7},  64827--64836 (2019)

\bibitem{ghimire2013geometric}
Ghimire, D., Lee, J.: Geometric feature-based facial expression recognition in
  image sequences using multi-class adaboost and support vector machines.
  Sensors  \textbf{13}(6),  7714--7734 (2013)

\bibitem{fer2013}
Goodfellow, I.J., Erhan, D., Carrier, P.L., Courville, A., Mirza, M., Hamner,
  B., Cukierski, W., Tang, Y., Thaler, D., Lee, D.H., et~al.: Challenges in
  representation learning: A report on three machine learning contests. In:
  International conference on neural information processing. pp. 117--124.
  Springer (2013)

\bibitem{happy2012real}
Happy, S., George, A., Routray, A.: A real time facial expression
  classification system using local binary patterns. In: 2012 4th International
  conference on intelligent human computer interaction (IHCI). pp.~1--5. IEEE
  (2012)

\bibitem{hasani2017facial}
Hasani, B., Mahoor, M.H.: Facial expression recognition using enhanced deep 3d
  convolutional neural networks. In: Proceedings of the IEEE CVPR workshops.
  pp. 30--40 (2017)

\bibitem{he2015deep}
He, K., Zhang, X., Ren, S., Sun, J.: Deep residual learning for image
  recognition. corr abs/1512.03385 (2015) (2015)

\bibitem{hu2018squeeze}
Hu, J., Shen, L., Sun, G.: Squeeze-and-excitation networks. In: Proceedings of
  the IEEE CVPR. pp. 7132--7141 (2018)

\bibitem{jeong2020deep}
Jeong, D., Kim, B.G., Dong, S.Y.: Deep joint spatiotemporal network (djstn) for
  efficient facial expression recognition. Sensors  \textbf{20}(7), ~1936
  (2020)

\bibitem{Jung_2015_ICCV}
Jung, H., Lee, S., Yim, J., Park, S., Kim, J.: Joint fine-tuning in deep neural
  networks for facial expression recognition. In: Proceedings of the IEEE
  International Conference on Computer Vision (ICCV) (December 2015)

\bibitem{kegel2020dynamic}
Kegel, L.C., Brugger, P., Fr{\"u}hholz, S., Grunwald, T., Hilfiker, P., Kohnen,
  O., Loertscher, M.L., Mersch, D., Rey, A., Sollfrank, T., et~al.: Dynamic
  human and avatar facial expressions elicit differential brain responses.
  Social cognitive and affective neuroscience  \textbf{15}(3),  303--317 (2020)

\bibitem{kimfer}
Kim, B.K., Roh, J., Dong, S.Y., Lee, S.Y.: Hierarchical committee of deep
  convolutional neural networks for robust facial expression recognition.
  Journal on Multimodal User Interfaces  \textbf{10}(2),  173--189 (2016)

\bibitem{kingma2014adam}
Kingma, D.P., Ba, J.: Adam: A method for stochastic optimization.
  arXiv:1412.6980  (2014)

\bibitem{kollias2017recognition}
Kollias, D., Nicolaou, M.A., Kotsia, I., Zhao, G., Zafeiriou, S.: Recognition
  of affect in the wild using deep neural networks. In: Proceedings of the IEEE
  CVPR Workshops. pp. 26--33 (2017)

\bibitem{kollias2020analysing}
Kollias, D., Schulc, A., Hajiyev, E., Zafeiriou, S.: Analysing affective
  behavior in the first abaw 2020 competition. arXiv:2001.11409  (2020)

\bibitem{affwild2}
Kollias, D., Zafeiriou, S.: Aff-wild2: Extending the aff-wild database for
  affect recognition. arXiv:1811.07770  (2018)

\bibitem{kollias2018multi}
Kollias, D., Zafeiriou, S.: A multi-task learning \& generation framework:
  Valence-arousal, action units \& primary expressions. arXiv:1811.07771
  (2018)

\bibitem{kollias2018training}
Kollias, D., Zafeiriou, S.: Training deep neural networks with different
  datasets in-the-wild: The emotion recognition paradigm. In: 2018
  International Joint Conference on Neural Networks (IJCNN). pp.~1--8. IEEE
  (2018)

\bibitem{kollias2019expression}
Kollias, D., Zafeiriou, S.: Expression, affect, action unit recognition:
  Aff-wild2, multi-task learning and arcface. arXiv:1910.04855  (2019)

\bibitem{kotsia2008analysis}
Kotsia, I., Buciu, I., Pitas, I.: An analysis of facial expression recognition
  under partial facial image occlusion. Image and Vision Computing
  \textbf{26}(7),  1052--1067 (2008)

\bibitem{kotsia2006facial}
Kotsia, I., Pitas, I.: Facial expression recognition in image sequences using
  geometric deformation features and support vector machines. IEEE transactions
  on image processing  \textbf{16}(1),  172--187 (2006)

\bibitem{li2021efficient}
Li, M., Li, X., Sun, W., Wang, X., Wang, S.: Efficient convolutional neural
  network with multi-kernel enhancement features for real-time facial
  expression recognition. Journal of Real-Time Image Processing pp. 1--12
  (2021)

\bibitem{rafdb}
Li, S., Deng, W., Du, J.: Reliable crowdsourcing and deep locality-preserving
  learning for expression recognition in the wild. In: Proceedings of the IEEE
  CVPR. pp. 2852--2861 (2017)

\bibitem{liu2020effective}
Liu, B., Ait-Boudaoud, D.: Effective image super resolution via hierarchical
  convolutional neural network. Neurocomputing  \textbf{374},  109--116 (2020)

\bibitem{fer2013result}
Luan, P., Huynh, V., Tuan~Anh, T.: Facial expression recognition using residual
  masking network. In: IEEE 25th International Conference on Pattern
  Recognition. pp. 4513--4519 (2020)

\bibitem{ck}
Lucey, P., Cohn, J.F., Kanade, T., Saragih, J., Ambadar, Z., Matthews, I.: The
  extended cohn-kanade dataset (ck+): A complete dataset for action unit and
  emotion-specified expression. In: 2010 ieee computer society CVPR-workshops.
  pp. 94--101. IEEE (2010)

\bibitem{ma2021robust}
Ma, F., Sun, B., Li, S.: Robust facial expression recognition with
  convolutional visual transformers. arXiv:2103.16854  (2021)

\bibitem{maaten2008visualizing}
Maaten, L.v.d., Hinton, G.: Visualizing data using t-sne. Journal of machine
  learning research  \textbf{9}(Nov),  2579--2605 (2008)

\bibitem{mahmoudiexp2}
Mahmoudi, M.A., Chetouani, A., Boufera, F., Tabia, H.: Learnable pooling
  weights for facial expression recognition. Pattern Recognition Letters
  \textbf{138},  644--650 (2020)

\bibitem{massoli2020cross}
Massoli, F.V., Amato, G., Falchi, F.: Cross-resolution learning for face
  recognition. Image and Vision Computing  \textbf{99},  103927 (2020)

\bibitem{michel2003real}
Michel, P., El~Kaliouby, R.: Real time facial expression recognition in video
  using support vector machines. In: Proceedings of the 5th international
  conference on Multimodal interfaces. pp. 258--264 (2003)

\bibitem{Oulu-CASIAresult}
Ming, Z., Xia, J., Luqman, M.M., Burie, J.C., Zhao, K.: Dynamic multi-task
  learning for face recognition with facial expression. arXiv:1911.03281
  (2019)

\bibitem{muhammad2017facial}
Muhammad, G., Alsulaiman, M., Amin, S.U., Ghoneim, A., Alhamid, M.F.: A
  facial-expression monitoring system for improved healthcare in smart cities.
  IEEE Access  \textbf{5},  10871--10881 (2017)

\bibitem{mmi}
Pantic, M., Valstar, M., Rademaker, R., Maat, L.: Web-based database for facial
  expression analysis. In: 2005 IEEE international conference on multimedia and
  Expo. pp. 5--pp. IEEE (2005)

\bibitem{Rouast_2019}
Rouast, P.V., Adam, M., Chiong, R.: Deep learning for human affect recognition:
  Insights and new developments. IEEE Transactions on Affective Computing p.
  1–1 (2019). \doi{10.1109/taffc.2018.2890471},
  \url{http://dx.doi.org/10.1109/TAFFC.2018.2890471}

\bibitem{russell1978evidence}
Russell, J.A.: Evidence of convergent validity on the dimensions of affect.
  Journal of personality and social psychology  \textbf{36}(10), ~1152 (1978)

\bibitem{russell1980circumplex}
Russell, J.A.: A circumplex model of affect. Journal of personality and social
  psychology  \textbf{39}(6), ~1161 (1980)

\bibitem{sajedi2020uncertainty}
Sajedi, S.O., Liang, X.: Uncertainty-assisted deep vision structural health
  monitoring. Computer-Aided Civil and Infrastructure Engineering  (2020)

\bibitem{shao2021fcnn}
Shao, J., Cheng, Q.: E-fcnn for tiny facial expression recognition. Applied
  Intelligence  \textbf{51}(1),  549--559 (2021)

\bibitem{shi2020human}
Shi, Y., Zhang, Z., Huang, K., Ma, W., Tu, S.: Human-computer interaction based
  on face feature localization. Journal of Visual Communication and Image
  Representation  \textbf{70},  102740 (2020)

\bibitem{suk2014real}
Suk, M., Prabhakaran, B.: Real-time mobile facial expression recognition
  system-a case study. In: Proceedings of the IEEE CVPR workshops. pp. 132--137
  (2014)

\bibitem{tang2013deep}
Tang, Y.: Deep learning using linear support vector machines. Proc. ICML
  Workshop Challenges Represent. Learn. Workshop p. 1–6 (2013)

\bibitem{cnn_2}
Tolias, G., Sicre, R., J{\'e}gou, H.: Particular object retrieval with integral
  max-pooling of cnn activations. arXiv:1511.05879  (2015)

\bibitem{class_retrieval}
Torresani, L., Szummer, M., Fitzgibbon, A.: Efficient object category
  recognition using classemes. In: European conference on computer vision. pp.
  776--789. Springer (2010)

\bibitem{raf-dbresultvo}
Vo, T.H., Lee, G.S., Yang, H.J., Kim, S.H.: Pyramid with super resolution for
  in-the-wild facial expression recognition. IEEE Access  \textbf{8},
  131988--132001 (2020)

\bibitem{wang2020regionraf}
Wang, K., Peng, X., Yang, J., Meng, D., Qiao, Y.: Region attention networks for
  pose and occlusion robust facial expression recognition. IEEE Transactions on
  Image Processing  \textbf{29},  4057--4069 (2020)

\bibitem{whissell1989dictionary}
Whissell, C.: The dictionary of affect in language, emotion: Theory, research
  and experience. the measurement of emotions, r. plutchik and h. kellerman,
  eds., vol. 4 (1989)

\bibitem{yang2017facialoulu}
Yang, B., Cao, J., Ni, R., Zhang, Y.: Facial expression recognition using
  weighted mixture deep neural network based on double-channel facial images.
  IEEE Access  \textbf{6},  4630--4640 (2017)

\bibitem{yolcu2020deep}
Yolcu, G., Oztel, I., Kazan, S., Oz, C., Bunyak, F.: Deep learning-based face
  analysis system for monitoring customer interest. Journal of Ambient
  Intelligence and Humanized Computing  \textbf{11}(1),  237--248 (2020)

\bibitem{yuoulu}
Yu, M., Zheng, H., Peng, Z., Dong, J., Du, H.: Facial expression recognition
  based on a multi-task global-local network. Pattern Recognition Letters
  \textbf{131},  166--171 (2020)

\bibitem{yu2015image}
Yu, Z., Zhang, C.: Image based static facial expression recognition with
  multiple deep network learning. In: Proceedings of the 2015 ACM on
  international conference on multimodal interaction. pp. 435--442 (2015)

\bibitem{zafeiriou2017aff}
Zafeiriou, S., Kollias, D., Nicolaou, M.A., Papaioannou, A., Zhao, G., Kotsia,
  I.: Aff-wild: valence and arousal'in-the-wild'challenge. In: Proceedings of
  the IEEE CVPR workshops. pp. 34--41 (2017)

\bibitem{zhang2020oulu}
Zhang, H., Huang, B., Tian, G.: Facial expression recognition based on deep
  convolution long short-term memory networks of double-channel weighted
  mixture. Pattern Recognition Letters  \textbf{131},  128--134 (2020)

\bibitem{zhang2017facialoulu}
Zhang, K., Huang, Y., Du, Y., Wang, L.: Facial expression recognition based on
  deep evolutional spatial-temporal networks. IEEE Transactions on Image
  Processing  \textbf{26}(9),  4193--4203 (2017)

\bibitem{oulucasia}
Zhao, G., Huang, X., Taini, M., Li, S.Z., Pietik{\"a}Inen, M.: Facial
  expression recognition from near-infrared videos. Image and Vision Computing
  \textbf{29}(9),  607--619 (2011)

\bibitem{zhao2016peakoulu}
Zhao, X., Liang, X., Liu, L., Li, T., Han, Y., Vasconcelos, N., Yan, S.:
  Peak-piloted deep network for facial expression recognition. In: European
  conference on computer vision. pp. 425--442. Springer (2016)

\bibitem{zhao2011facial}
Zhao, X., Zhang, S.: Facial expression recognition based on local binary
  patterns and kernel discriminant isomap. Sensors  \textbf{11}(10),
  9573--9588 (2011)

\end{thebibliography}
